\documentclass[a4paper,12pt]{article}

\usepackage{orcidlink,thumbpdf,lmodern}
\usepackage{amsmath}
\usepackage{amsthm}
\usepackage{amsfonts}
\usepackage[hypcap=false]{caption}
\usepackage{listings}
\usepackage{lscape}
\usepackage{natbib}
\usepackage{graphicx}
\usepackage{authblk}
\usepackage{geometry} 
\usepackage{graphicx}%
\usepackage{multirow}%
\usepackage{mathrsfs}%
\usepackage[title]{appendix}%
\usepackage{xcolor}%
\usepackage{textcomp}%
\usepackage{manyfoot}%
\usepackage{booktabs}%
\usepackage{algorithm}%
\usepackage{algorithmicx}%
\usepackage{algpseudocode}%
\usepackage[export]{adjustbox}
\usepackage{pdfpages}

\theoremstyle{plain}
\newtheorem{proposition}{Proposition}[section]
\newtheorem{lemma}{Lemma}[section]
\newtheorem{theorem}{Theorem}[section]
\theoremstyle{definition}
\newtheorem{definition}{Definition}[section]
\newtheorem{remark}{Remark}[section]

\newcommand{\E}{{\mathbb{E}}}
\newcommand{\IF}{\operatorname{IF}}
\newcommand{\HD}{\operatorname{HD}}
\newcommand{\thetaz}{{\theta_0}}
\newcommand{\alphaz}{{\alpha_0}}
\newcommand{\thetae}{{\theta_\varepsilon}}
\newcommand{\alphae}{{\alpha_\varepsilon}}

\newcommand{\dx}{d \mu(x)}

\begin{document}

\title{Hellinger loss function for Generative Adversarial Networks}

\author[1]{Giovanni Saraceno\thanks{\href{mailto:giovanni.saraceno@unipd.it}{giovanni.saraceno@unipd.it}}\orcidlink{0000-0002-1753-2367}}
\author[2]{Anand N. Vidyashankar}
\author[3]{Claudio Agostinelli}

\affil[1]{Department of Statistical Sciences, University of Padova, Italy}
\affil[2]{Department of Statistics, George Mason University, VA, USA}
\affil[3]{Department of Mathematics, University of Trento, Italy}

\date{}

\maketitle

\begin{abstract}
We propose Hellinger-type loss functions for training Generative Adversarial Networks (GANs), motivated by the boundedness, symmetry, and robustness properties of the Hellinger distance. We define an adversarial objective based on this divergence and study its statistical properties within a general parametric framework. We establish the existence, uniqueness, consistency, and joint asymptotic normality of the estimators obtained from the adversarial training procedure. In particular, we analyze the joint estimation of both generator and discriminator parameters, offering a comprehensive asymptotic characterization of the resulting estimators.
We introduce two implementations of the Hellinger-type loss and we evaluate their empirical behavior in comparison with the classic (Maximum Likelihood-type) GAN loss. Through a controlled simulation study, we demonstrate that both proposed losses yield improved estimation accuracy and robustness under increasing levels of data contamination. 

\smallskip 
    
\noindent \textbf{Keywords}:Generative models, Generative Adversarial Networks, Hellinger distance, Outliers, Robustness.

\end{abstract}

\section{Introduction}
\label{sec:introduction}

Deep learning models have been widely used across a variety of machine learning problems achieving great advances and have received increasing attention in data science and statistics. In fact, deep neural networks can be viewed as a non-linear and highly-parametrized generalization of statistical models \citep{Yuan2020}. 
One of the tasks solved by deep neural networks is called generative modeling, in which we are interested in learning a model capable of describing the underlying probability distribution given a sample of data. With the learned model, we are able to generate new data. 
More recently proposed generative models proceed by an adversarial procedure, based on the idea that a data generator is good if generated data, labeled as ``fake'', cannot be distinguished from real data. 
Generative Adversarial Networks (GANs), introduced by \citet{Goodfellow2014}, are considered state of the art for generative models and have been developed in several fields from both practical applications and theoretical analysis. 
The purpose of GANs is to generate observations that are similar to samples collected by a target distribution $p_\ast$. GANs are conducted by an adversarial procedure that involves a family of generators and a family of discriminators, usually implemented by neural networks.
In particular, the two networks are trained together in a minimax game: the generators transform low-dimensional observations drawn from a known density (usually normal or uniform) into fake observations trying to imitate $p_\ast$, while the goal of discriminators is to accurately discriminate between the samples from $p_\ast$ and the generated ``fake'' data. GANs have been successfully applied in various domains, ranging from computer vision to natural language processing and medical imaging, achieving state-of-the-art performance.

Significant effort has been devoted to studying GANs from a statistical and theoretical perspective. 
The early theoretical work by \citet{Biau2020} provides an initial analysis of the asymptotic properties of the GAN estimators, showing that the original formulation is linked to the Jensen--Shannon divergence and establishing convergence results under regularity assumptions and smoothness conditions.
Despite their success, GANs are known to be challenging to train, suffering issues like unstable convergence, vanishing gradients, and mode collapse. To address these problems, the original formulation has been extended by employing alternative divergence measures. 
In particular, \citep{Nowozin2016} introduce the $f$GAN framework in which any $f$-divergence can be used as a training objective by way of a variational formulation, and \citep{arjovsky2017} consider the Wasserstein distance by introducing the Wasserstein GAN (WGAN), which is shown to improve training stability and mitigate vanishing gradients. %, replacing the binary classification-based discriminator with a critic that estimates the Wasserstein-1 distance between the real and generated distributions. 
These developments underscored that the choice of the divergence considered may be critical to GAN performance. Following this thought, several variants of GAN have been proposed, grounded in different statistical distances, e.g., the Least-Square GANs \citep{mao2017least} and $W_2$-GAN \citep{korotin2019wasserstein}. Recent surveys, such as \citet{Chakraborty2024}, further report the large amount of GAN variants, including the Cumulant GANs, introduced by \citet{pantazis2023cumulant}  which replace classical divergences with a framework based on cumulant generating functions, offering theoretical connections to Rényi divergences and improved gradient properties during training, and Relativistic GAN, that aim to address training instability and mode collapse by modifying the loss function or discriminator architecture.

At the same time, there is a growing interest in establishing a theoretical understanding of GANs. For example, following the initial asymptotic analysis of GAN estimators by \cite{Biau2020}, \citet{chakraborty2024-low} have explored the generalization behavior and statistical efficiency of GANs in regimes where data lie on low-dimensional structures embedded in high-dimensional spaces. This is inspired by real-world data, such as images or sensory signals, which often possess a low intrinsic dimensionality despite their high ambient representation. \citet{chakraborty2024-low} provide rigorous convergence rate analyses for both GANs and their bidirectional variants (BiGANs). 
Collectively, these developments reflect an ongoing movement in the GAN literature toward frameworks that are not only efficient in practice but also with a strong theoretical basis. 

Recent research has also focused on adapting adversarial training frameworks to enhance robustness against data contamination. Classical robust statistics suggest that using bounded divergence measures can yield estimators resistant to the influence of outliers. For example, \citet{gao2019robust} analyze the relationship between GANs and classical depth-based estimators, showing that the adversarial formulations can achieve optimal rates in robust location and scatter estimation problems when the discriminator is properly constructed. \citet{GaoYaoZhu2020} similarly study robust covariance matrix estimation through proper scoring rules, induced by variational approximation of $f$-divergences. \citet{ZhuJiaoJordan2022} propose a general theoretical framework for GAN-based estimators of unknown parameters of the true distribution that satisfy robustness guarantees under broad distributional conditions, including sub-exponential classes. 
\cite{Zhang2023} introduce a robust GAN framework that trains the generator and discriminator against worst-case perturbations. \cite{azimi2024zganoutlierfocusedgenerativeadversarial} develop zGAN, an outlier-focused GAN for synthetic data, which explicitly generates realistic outliers and tail events to augment training data.
While these studies indicate the potential of GANs in robust inference, they often consider specific aspects (e.g. robust loss design or rate optimality) in isolation. From this point of view, two key gaps can be identified. First, most theoretical analyses of GANs treat the generator and discriminator separately, for instance, assuming an optimal discriminator and focusing on the asymptotic properties of the generator. Second, the robustness of GAN estimators is rarely examined using statistical tools such as the influence function and the resistance to outliers.  

In this work, we propose a Hellinger-type loss function for GAN training and investigate the theoretical properties of the corresponding estimators. The Hellinger distance is a symmetric, bounded divergence between two density functions with a long history in statistics and well known connection to robust estimation. Our contributions can be summarized as follows. We define a novel adversarial objective based on a Hellinger-type distance, aiming to reduce the influence of outliers on the estimators. We develop a comprehensive asymptotic theory analyzing the generator and discriminator parameters jointly. In particular, we establish the existence and uniqueness of the estimators, prove consistency, and derive their joint asymptotic normality. We investigate the robustness of the proposed Hellinger GAN estimator. Specifically, we derive the influence function of the joint estimator, which provides insights into its sensitivity to model contamination. 

The remainder of the paper is organized as follows. Section~\ref{sec:hellinger-loss} formally introduces the Hellinger-type loss function in the context of adversarial training and define the associated optimization objective. Section~\ref{sec:asymtotic} presents the main theoretical results, including the existence, uniqueness, consistency, and joint asymptotic normality of the estimator under appropriate regularity conditions. Section~\ref{sec:IF} examines the robustness properties by deriving the influence functions. Section~\ref{sec:simulation} contains the numerical experiments in a Gaussian setting in which we evaluate the performance of the Hellinger-type GAN loss in the presence of contamination. Section~\ref{sec:fashionMNIST} presents results on the Fashion-MNIST dataset. Finally, Section~\ref{sec:conclusion} concludes the paper.

\section{Hellinger-type Loss}
\label{sec:hellinger-loss}

From a mathematical point of view, we can represent the process of GANs as follows. 
Let $X_1, \ldots, X_n$ be i.i.d. observations sampled from some unknown density $p_\ast$ on $E$, where $E$ is a Borel subset of $\mathbb{R}^d$. The density $p_\ast$ is supposed to be dominated by a fixed known measure $\mu$ on $E$ and this condition holds for all densities we consider here. Let $Z$ be a $d^\prime$-random variable with density $g$, where $d^\prime \ll d$. The generators can be represented by a parametric family of functions from $\mathbb{R}^{d^\prime}$ to $E$, that is, $\mathcal{G} = \{G_\theta\}_{\theta \in \Theta}$, $\Theta \subset \mathbb{R}^p$. Each function $G_\theta$ is applied to the variable $z$, which is usually called latent variable or noise, so that we can consider the natural family of density $\mathcal{Q} = \{q_\theta\}_{\theta \in \Theta}$ associated with the generators defined as $G_\theta(Z) \stackrel{\mathcal{L}}{=} q_\theta d\mu$, where these densities are possible candidates to represent $p_\ast$. On the other hand, the family of discriminators $\mathcal{D} = \{D_\alpha\}_{\alpha \in \Lambda}$, $\Lambda \subset \mathbb{R}^q$, can be described by a family of functions from $E$ to $[0,1]$. The value computed by the discriminator can be thought of as the probability that a given observation comes from the true density $p_\ast$. Notice that, since generators and discriminators are usually represented by neural networks, the dimensions $p$ and $q$ can be very large. 

Let $Z_1, \ldots, Z_m$ be an i.i.d. sample distributed as the latent variable $Z\sim g$. According to the standard formulation of GANs, discriminators and generators are fine-tuned by optimizing the objective function 
\begin{equation*}
    L_n(\theta,\alpha) = \frac{1}{n}\sum_{i=1}^n \ln(D_\alpha(X_i)) + \frac{1}{m}\sum_{j=1}^m \ln(1-D_\alpha ( G_\theta(Z_j)))
\end{equation*}
with respect to $(\theta, \alpha)$, where $\ln$ indicates the natural logarithm. The corresponding population version is given by 
\begin{equation}
    \label{eq:classical_loss}
    L(\theta,\alpha) = \int \ln(D_\alpha(x)) p_\ast(x)d\mu(x) + \int \ln(1-D_\alpha( G_\theta(z))) g(z) d\mu(z).
\end{equation}
Therefore, this objective function represents the adversarial game between discriminators and generators: for a given $\theta$, the discriminator is determined to be minimal in generated data $G_\theta(Z_j)$, $j = 1, \ldots,m$, and maximal on samples $X_i$, $i = 1, \ldots, n$; on the other hand, for a given $\alpha$, the generator is chosen so that $D_\alpha (G_\theta(Z_j))$ are maximized. Hence, we want to find $(\hat{\theta}, \hat{\alpha})$ such that
\begin{equation}
  \label{eq:hat-L}
    (\hat{\theta}, \hat{\alpha}) = \arg \inf_{\theta \in \Theta} \sup_{\alpha \in \Lambda} L_n(\theta,\alpha).
\end{equation}
\citet{Goodfellow2014} showed that the objective function given in equation (\ref{eq:hat-L}), under appropriate conditions, reduces to the Jensen-Shannon divergence between the data generating density $p_\ast$ and the family of parameterized densities. Following the idea of the connection between the loss function and a divergence, we propose to use a loss function constructed by using the Hellinger distance. 

\begin{definition}
	Consider the measurable space $(\mathcal{X,F})$ and a $\sigma$-finite measure $\lambda$ on $(\mathcal{X,F})$. For every $u, v \in L^2(\mathcal{X})$ such that $u$ and $v$ are dominated by the measure $\lambda$, the squared Hellinger distance between $u$ and $v$ is given by
    \begin{equation*}
  	d_{HD}(u,v) = \int (u^{\frac{1}{2}} - v^{\frac{1}{2}})^2 d\lambda.
	\end{equation*}
\end{definition}
Notice that the Hellinger distance can be rewritten as
\begin{equation*}
  d_{HD}(u,v) = \int u  \mbox{ } d\lambda + \int v \mbox{ } d\lambda - 2\int |u v|^{\frac{1}{2}} d\lambda. 
\end{equation*}
As special case, if $u$ and $v$ are density probability functions it simplifies to
\begin{equation*}
  d_{HD}(u,v) = 2 - 2\int |u v|^{\frac{1}{2}} d\lambda.
\end{equation*}
Considering our setting, we want to construct a loss function such that the Hellinger distance between the functions $D_\alpha$ and $(1 - D_\alpha( G_\theta))$, for $\theta \in \Theta$ and $\alpha \in \Lambda$, is optimized. Defining $f(x,z) = p_\ast(x) g(z)$ and $\mu(x,z) = \mu(x) \times \mu(z)$, we can consider $d\lambda(x,z) = f(x,z) d\mu(x,z)$. Hence, the Hellinger distance between the functions $D_\alpha$ and $(1 - D_\alpha (G_\theta))$ is given by
\begin{align*}
  HD^2(D_\alpha,1-D_\alpha (G_\theta)) = & \int (D_\alpha^{\frac{1}{2}}(x) - (1-D_\alpha (G_\theta(z)))^{\frac{1}{2}})^2 d\lambda(x,z) \\
  = & \int D_\alpha(x)  d\lambda(x,z) + \int (1-D_\alpha  (G_\theta(z))) d\lambda(x,z) \\
  & - 2\int D^{\frac{1}{2}}_\alpha(x)(1-D_\alpha  (G_\theta(z)))^{\frac{1}{2}} d\lambda(x,z) .
\end{align*}
In particular, since $f$ is a product density and $\mu$ is a product measure we have $\int D_\alpha(x)  d\lambda(x,z) = \int D_\alpha(x)p_\ast(x) d\mu(x)$ and $\int (1-D_\alpha(G_\theta(z))) d\lambda(x,z) = \int (1-D_\alpha (G_\theta(z))) g(z) d\mu(z)$. Hence, the objective function has the form
\begin{align}
  \nonumber HD^2(\theta, \alpha) = & \int D_\alpha(x)p_\ast(x)  d\mu(x)   + \int (1-D_\alpha (G_\theta(z))) g(z) d\mu(z) \\
  \nonumber & - 2 \gamma(\theta,\alpha) \\
  = & h_1(\alpha) + h_2(\theta, \alpha) - 2 \gamma(\theta,\alpha), 
  \label{eq:hd-complete}
\end{align}
where
\begin{equation}
	\label{eq:obj-gamma}
	\gamma(\theta,\alpha) = \int D^{\frac{1}{2}}_\alpha(x) p_\ast(x) d\mu(x)\int (1-D_\alpha(G_\theta(z)))^{\frac{1}{2}}  g(z) d\mu(z).
\end{equation}
We will denote $\mathrm{HD}^2_n(\theta, \alpha)$ and $\gamma_n(\theta, \alpha)$ the corresponding sample versions. 

\subsection{Approximated objective function}
\label{sec:approx-obj-fun}
 
Consider the same setting presented in the previous section. Notice that, the objective function given in (\ref{eq:classical_loss}) can be rewritten as
\begin{align*}
	L(\theta,\alpha) = & \int \ln(D_\alpha(x))p_\ast(x) d\mu(x) + \int \ln(1-D_\alpha(G_\theta(z)))g(z)d\mu(z) \\
	=& \int \ln(D_\alpha(x))p_\ast(x) d\mu(x) \int g(z)d\mu(z) + \\
	&   \int \ln(1-D_\alpha \circ G_\theta(z))g(z)d\mu(z) \int p_\ast(x)d\mu(x) \\
	=& 2 \int \ln(D_\alpha(x)(1-D_\alpha (G_\theta(z))))^{\frac{1}{2}} f(x,z) d\mu(x,z) \\
	=& 2 \int \ln(1 + (D^{\frac{1}{2}}_\alpha(x)(1-D_\alpha (G_\theta(z)))^{\frac{1}{2}} - 1)) d\lambda(x,z) . \\
\end{align*}
Therefore, by first order Taylor's series approximation, the function $L(\theta,\alpha)$ is approximated by the objective function
\begin{align*}
	\label{eq:obj-approx}
\tilde{L}(\theta,\alpha) & = -2 \left( 1 - \int D^{\frac{1}{2}}_\alpha(x) (1-D_\alpha(G_\theta(z)))^{\frac{1}{2}} d\lambda(x,z) \right) \\
& = 2 \gamma(\theta,\alpha) - 2 .
\end{align*}
and hence $\tilde{L}_n(\theta,\alpha) = 2 \gamma_n(\theta,\alpha) - 2$ is an approximation of $L_n(\theta,\alpha)$.
Notice that the objective function $\tilde{L}(\theta,\alpha)$ resembles the formulation of Hellinger distance between two density functions of equations (\ref{eq:hd-complete}) and (\ref{eq:obj-gamma}). In particular, the first two terms are referred to well-separated parts of the GAN process: while the first term is related to how the discriminators evaluate the data, the second term does not depend on data but only on generated data from $Z$. 

Finally, we are interested in finding $(\theta_n, \alpha_n)$ such that
\begin{equation}
	\label{eq:hat-H}
	(\theta_n, \alpha_n) = \arg \inf_{\theta \in \Theta} \sup_{\alpha \in \Lambda} HD_n^2(\theta,\alpha),
\end{equation}
or
\begin{align}
	\label{eq:hat-gamma}
	(\theta_n, \alpha_n) & = \arg \inf_{\theta \in \Theta} \sup_{\alpha \in \Lambda} \tilde{L}_n(\theta,\alpha) \nonumber \\
  & = \arg \inf_{\theta \in \Theta} \sup_{\alpha \in \Lambda} \gamma_n(\theta,\alpha).
\end{align}

\subsection{Divergence-based Losses}
\label{subsec:divergence}

The original GAN \citep{Goodfellow2014}, as described in (\ref{eq:classical_loss}), solves
\begin{equation*}
    \inf_\theta \sup_\alpha \Big\{ \E_{X\sim p_*}[\ln D_\alpha(X)] + \E_{Z\sim g}[\ln(1-D_\alpha(G_\theta(Z)))] \Big\}.
\end{equation*}
Under the optimal discriminator 
$D^\ast_\theta(x)=p_*(x)/\big(p_*(x)+q_\theta(x)\big)$, this reduces to minimizing the symmetric Jensen--Shannon divergence ($\mathrm{JSD}$) between the data distribution $p_\ast$ and the family of parametrized densities $q_\theta$. The JSD is bounded and symmetric, but it can saturate, leading to vanishing gradients for the generator. 
The Wasserstein GAN (WGAN; \citealp{arjovsky2017}) replaces the JSD objective by minimizing the 1-Wasserstein distance, that is
\begin{equation*}
    \inf_\theta \sup_{D:\,\|D\|_L \leq 1} \Big\{ \E_{X\sim p_*}[D(X)] - \E_{Z\sim g}[D(G_\theta(Z))] \Big\},
\end{equation*}
where the supremum is over all 1-Lipschitz functions $D$.  By the Kantorovich--Rubinstein duality this maximization computes the Wasserstein metric enforcing a Lipschitz gradient constraint on $D_\alpha$, and importantly WGANs yield non-vanishing gradients.
More generally, \cite{Nowozin2016} ($f$-GAN) show that any $f$-divergence can be employed by introducing a variational discriminator $T_\alpha$ and its Fenchel--Legendre dual $f^\ast$. The $f$-GAN framework solves
\begin{equation*}
    \inf_\theta \sup_T \Big\{ \E_{X\sim p_\ast}[T(X)] - \E_{Z\sim g}[f^\ast(T(G_\theta(Z)))] \Big\},
\end{equation*}
which under optimal $T_\alpha$ is equivalent to minimizing the chosen $f$-divergence between $p_\ast$ and $q_\theta$. 
 Different choices of $f$ include the Kullback--Leibler, Pearson $\chi^2$, JSD, and the squared Hellinger distance. The stability and robustness properties depend strongly on the choice of $f$.
 
\noindent 
The proposed Hellinger-GAN instead directly targets the squared Hellinger distance between the discriminator on data and the discriminator on generated data, that is
\begin{equation*}
    \inf_{\theta \in \Theta} \sup_{\alpha \in \Lambda} \mathrm{HD}^2(\theta, \alpha) = \inf_{\theta \in \Theta} \sup_{\alpha \in \Lambda} \; \mathrm{HD}^2(D_\alpha(\cdot),1-D_\alpha(G_\theta(\cdot))).
\end{equation*}
The squared Hellinger distance is symmetric and bounded%($0 \le HD^2 \le 2$)
, and its square root form may help to maintain the gradients globally finite. 
 
In summary, divergence-based GANs vary in whether their losses are bounded, symmetric, and gradient-stable. Additionally, the existing literature develops asymptotic guaranties only for the generator parameters under optimal discriminator parameters. In our framework, we investigate the joint asymptotic behavior of both the generator and the discriminator parameters. The details of this joint analysis are developed in the following sections.

\section{Asymptotic Properties}
\label{sec:asymtotic}
In this section we discuss existence and uniqueness of the proposed estimators together with their statistical asymptotic properties such as consistency and asymptotic normality of the estimators.

\subsection{Existence and Uniqueness}

We are interested in studying the properties of the objective function $\mathrm{HD}^2(\theta,\alpha)$ solutions with respect to $(\theta,\alpha)$, which is equivalent to find the couple of parameters $(\theta_\ast,\alpha_\ast)$ such that
\begin{equation}
\label{eqn:solution-star}
  (\theta_\ast,\alpha_\ast) = \arg \inf_{\theta \in \Theta} \sup_{\alpha \in \Lambda} \mathrm{HD}^2(\theta,\alpha).
\end{equation}
Our first result concerns the existence and uniqueness of $(\theta_\ast,\alpha_\ast)$. 
Consider the following assumptions.
\begin{itemize}
    \item[($H_G$)]  $G_\theta$ is continuous with respect to $\theta$, i.e. if $\theta_n \to \theta$ as $n \to \infty$, then $G_{\theta_n} \to G_\theta$ as $n \to \infty$. $\Theta$ is a compact subset of $\mathbb{R}^p$ and the model $\{G_{\theta}\}_{\theta \in \Theta}$ is identifiable with respect to $\theta$. 
    \item[($H_D$)] the function $(x,\alpha) \to D_\alpha (x)$ is $\mathcal{C}^1$, i.e. continuous and differentiable, with continuous differential. $\Lambda$ is a compact subset of $\mathbb{R}^q$ and the model $\{D_{\alpha}\}_{\alpha \in \Lambda}$ is identifiable with respect to $\alpha$.
    
\end{itemize}

\begin{theorem}{(Existence and uniqueness)}
    If $(H_D)$ and $(H_G)$ hold, then there exists a unique $(\theta_\ast,\alpha_\ast)$ such that
\begin{equation*}
    (\theta_\ast,\alpha_\ast)=\arg\inf_{\theta\in\Theta} \sup_{\alpha\in\Lambda} \mathrm{HD}^2(\theta,\alpha).
\end{equation*}

\end{theorem}
\begin{proof}
To prove the existence, we need to show that $\mathrm{HD}^2(\theta,\alpha)$ is jointly continuous with respect to $\theta$ and $\alpha$, that is, given $\alpha, \alpha_n \in \Lambda$ and $\theta, \theta_n \in \Theta$ such that $(\theta_n,\alpha_n) \longrightarrow (\theta,\alpha)$, then
\begin{equation*}
    \mathrm{HD}^2(\theta_n,\alpha_n) \longrightarrow \mathrm{HD}^2(\theta,\alpha) \mbox{ as } n\to \infty.
\end{equation*}
Note that 
\begin{align*}
    \mathrm{HD}^2(\theta_n,\alpha_n) - \mathrm{HD}^2(\theta,\alpha) =& h_1(\alpha_n) - h_1(\alpha) + h_2(\theta_n,\alpha_n) - h_2(\theta,\alpha) - 2\gamma(\theta_n,\alpha_n) + 2\gamma(\theta,\alpha) \\
    =& H_1 + H_2 + H_3
\end{align*}
First
\begin{align*}
    H_1 = h_1(\alpha_n) - h_1(\alpha) &= \int D_{\alpha_n}(x) p_\ast(x) d\mu(x) -\int D_\alpha(x) p_\ast(x) d\mu(x)\\
    &= \int (D_{\alpha_n}(x)  - D_\alpha(x)) p_\ast(x) d\mu(x).
\end{align*}
Notice that $(D_{\alpha_n}(x)  - D_\alpha(x))\le 2$, then by the Dominated Convergence Theorem (DCT)
\begin{equation*}
    \lim_{n\to\infty}|H_1| \le \int \lim_{n\to\infty} |D_{\alpha_n}(x)  - D_\alpha(x)|p_\ast(x)d\mu(x).
\end{equation*}
This is equal to zero by the continuity of $D_\alpha(x)$ stated in assumption $(H_D)$. Considering the notation $D_{\alpha,\theta}(z) = D_\alpha(G_\theta(z))$,
\begin{align*}
    H_2 =& h_2(\theta_n,\alpha_n) - h_2(\theta,\alpha) \\
    =& \int (1-D_{\alpha_n,\theta_n} (z))g(z) d\mu(z) -\int (1-D_{\alpha,\theta} (z))g(z) d\mu(z)\\
    %=& \int [(1-D_{\alpha_n,\theta_n}(z)) - (1-D_{\alpha,\theta} (z))] g(z)d\mu(z)\\
    =& \int [(1-D_{\alpha_n,\theta_n} (z)) -(1-D_{\alpha_n,\theta} (z))]q_{\theta}(z) d\mu(z) \\
    & + \int [(1-D_{\alpha_n,\theta} (z)) -(1-D_{\alpha,\theta} (z))]q_{\theta}(z) d\mu(z).
\end{align*}
Since $D_{\alpha,\theta}(z) \le 1$, by the DCT we have 
\begin{align*}
    \lim_{n\to\infty}|H_2| \le&  \int \lim_{n\to\infty} |(1-D_{\alpha_n,\theta_n} (z)) - (1-D_{\alpha_n,\theta} (z))| g(z)d\mu(z)\\
    & + \int \lim_{n\to\infty}|(1-D_{\alpha_n,\theta} (z)) -(1-D_{\alpha,\theta} (z))|g(z) d\mu(z).
\end{align*}
The first term is equal to zero for the continuity of $D_{\alpha} (x)$ with respect to $x$ by ($H_D$) and the continuity of $G_{\theta} (z)$ by $(H_G)$. By the continuity of $D_{\alpha} (x)$ with respect to $\alpha$, also the second term is zero.

Finally, we prove that 
\begin{equation*}
    \lim_{n\to\infty} |H_3| = 2\lim_{n\to\infty} |\gamma(\theta_n,\alpha_n) - \gamma(\theta,\alpha)| =0.
\end{equation*}
%\allowdisplaybreaks
Notice that
\begin{align*}
    \gamma(\theta_n,\alpha_n) - \gamma(\theta,\alpha) %= & \int D_{\alpha_n}^{\frac{1}{2}}(x) p_\ast(x) d\mu(x) \int (1-D_{\alpha_n,\theta_n} (z))^{\frac{1}{2}}g(z) d\mu(z)\\
    %& - \int D_{\alpha}^{\frac{1}{2}}(x) p_\ast(x) d\mu(x) \int (1-D_{\alpha,\theta} (z))^{\frac{1}{2}}g(z) d\mu(z)\\
    =& \int D_{\alpha_n}^{\frac{1}{2}}(x) p_\ast(x) d\mu(x) \int \left[(1-D_{\alpha_n,\theta_n} (z))^{\frac{1}{2}} - (1-D_{\alpha_n,\theta} (z))^{\frac{1}{2}}\right] g(z) d\mu(z)\\
    & + \int D_{\alpha_n}^{\frac{1}{2}}(x) p_\ast(x) d\mu(x) \int (1-D_{\alpha_n,\theta} (z))^{\frac{1}{2}} g(z) d\mu(z)\\
    & - \int D_{\alpha}^{\frac{1}{2}}(x) p_\ast(x) d\mu(x) \int (1-D_{\alpha,\theta} (z))^{\frac{1}{2}}g(z) d\mu(z)\\
\end{align*}
where we added and subtracted the quantity $(1- D_{\alpha_n, \theta}(z))^{\frac{1}{2}}g(z)$ in the second integral. Repeating the same operation with $(1-D_{\alpha,\theta}(z))^{\frac{1}{2}} g(z)$ we get
\begin{align*}
\gamma(\theta_n,\alpha_n) - \gamma(\theta,\alpha) & = \int D_{\alpha_n}^{\frac{1}{2}}(x) p_\ast(x) d\mu(x) \int \left[(1-D_{\alpha_n,\theta_n} (z))^{\frac{1}{2}} - (1-D_{\alpha_n,\theta} (z))^{\frac{1}{2}}\right] g(z) d\mu(z)\\
    & + \int D_{\alpha_n}^{\frac{1}{2}}(x) p_\ast(x) d\mu(x) \int \left[(1-D_{\alpha_n,\theta} (z))^{\frac{1}{2}} - (1-D_{\alpha,\theta} (z))^{\frac{1}{2}}\right] g(z) d\mu(z)\\
    & + \int D_{\alpha_n}^{\frac{1}{2}}(x) p_\ast(x) d\mu(x) \int (1-D_{\alpha,\theta} (z))^{\frac{1}{2}} g(z) d\mu(z)\\
    & - \int D_{\alpha}^{\frac{1}{2}}(x) p_\ast(x) d\mu(x) \int (1-D_{\alpha,\theta} (z))^{\frac{1}{2}}g(z) d\mu(z).\\
\end{align*}
Finally, one more iteration with $D_{\alpha}^{\frac{1}{2}}(x) p_\ast(x)$ in the first integral leads to
\begin{align*}
    \gamma(\theta_n,\alpha_n) - \gamma(\theta,\alpha) = & \int D_{\alpha_n}^{\frac{1}{2}}(x) p_\ast(x) d\mu(x) \int \left[(1-D_{\alpha_n,\theta_n} (z))^{\frac{1}{2}} - (1-D_{\alpha_n,\theta} (z))^{\frac{1}{2}}\right] g(z) d\mu(z)\\
    & + \int D_{\alpha_n}^{\frac{1}{2}}(x) p_\ast(x) d\mu(x) \int \left[(1-D_{\alpha_n,\theta} (z))^{\frac{1}{2}} - (1-D_{\alpha,\theta} (z))^{\frac{1}{2}}\right] g(z) d\mu(z)\\
    & + \int \left( D_{\alpha_n}^{\frac{1}{2}}(x) -D_{\alpha}^{\frac{1}{2}}(x) \right) p_\ast(x) d\mu(x) \int (1-D_{\alpha,\theta} (z))^{\frac{1}{2}} g(z) d\mu(z)\\
    = & J_{1n} + J_{2n} + J_{3n}
\end{align*}
We consider the limit as $n \to +\infty$ of these integrals separately. Note that $\int D_{\alpha_n}^{\frac{1}{2}}(x) p_\ast(x) d\mu(x) \le 1$, $\forall n$, and $\left[ (1 - D_{\alpha_n, \theta_n}(z))^{\frac{1}{2}} - (1 - D_{\alpha_n, \theta}(z))^{\frac{1}{2}}\right] \le 2$. Then, for the DCT we have 
\begin{align*}
   \lim_{n \to +\infty } |J_{1n}| \le &  \quad \int \lim_{n \to +\infty} \left| (1 - D_{\alpha_n, \theta_n}(z))^{\frac{1}{2}} - (1 - D_{\alpha_n, \theta}(z))^{\frac{1}{2}}\right| g(z) d\mu(z)  
\end{align*}
 and this is equal to zero by the continuity of $G_\theta(z)$ with respect to $\theta$ and the continuity of $D_\alpha(x)$ with respect to $x$. Similarly, by the DCT 
\begin{equation*}
   \lim_{n \to +\infty } |J_{2n}| \le \int \lim_{n \to +\infty} \left|(1-D_{\alpha_n, \theta}(z))^{\frac{1}{2}} - (1-D_{\alpha, \theta}(z))^{\frac{1}{2}}\right| g(z) d\mu(z) = 0
\end{equation*}
by the continuity of $D_\alpha(x)$ with respect to $\alpha$.
Finally, note that in $J_{3n}$ the second integral does not depend on $n$, therefore it is not considered, and $\left ( D_{\alpha_n}^{\frac{1}{2}}(x) - D_{\alpha}^{\frac{1}{2}}(x)\right) \le 2$. Hence, for the DCT we have
\begin{equation*}
    \lim_{n \to +\infty } |J_{3n}|   \le  \int \lim_{n \to +\infty} \left| D_{\alpha_n}^{\frac{1}{2}} (x) - D_{\alpha}^{\frac{1}{2}} (x) \right| p_\ast(x) = 0. 
\end{equation*}
Considering the limit %of the supremum 
\begin{equation*}
    \lim_{n \to +\infty} |\gamma(\theta_n,\alpha_n) - \gamma(\theta,\alpha)| \le \lim_{n \to + \infty} (|J_{1n}| + |J_{2n}| + |J_{3n}|) = 0,
\end{equation*}
then
\begin{equation*}
    \lim_{n\to\infty} |HD^2(\theta_n,\alpha_n) - HD^2(\theta,\alpha)| \le \lim_{n\to\infty} (|H_1| + |H_2| + |H_3|) = 0.
\end{equation*}
We proved that the set $\{(\theta_\ast,\alpha_\ast) | (\theta_\ast,\alpha_\ast) = \arg \inf_{\theta \in \Theta} \sup_{\alpha \in \Lambda} \mathrm{HD}^2(\theta,\alpha)\}$ is not empty. It remains to prove the uniqueness. Assume that there exists $(\tilde{\theta}, \tilde{\alpha}) \in \Theta \times \Lambda$ such that
\begin{equation*}
    (\tilde{\theta},\tilde{\alpha})=\arg\inf_{\theta\in\Theta} \sup_{\alpha\in\Lambda} HD^2(\theta,\alpha).
\end{equation*}
This means that $HD^2(\tilde{\theta},\tilde{\alpha}) = HD^2(\theta_\ast,\alpha_\ast)$. Hence by the identifiability assumption, $(\tilde{\theta},\tilde{\alpha}) = (\theta_\ast,\alpha_\ast)$.
\end{proof}

\subsection{Consistency}

We now want to prove the consistency property. Let $d\mu_\ast(x) = p_\ast(x)d\mu(x)$ and $d\mu_g(z) = g(z)d\mu(z)$ be the probability measures induced by the density $p_\ast$ and $g$, respectively, and let $\mu_n(x)$ denote a sample-based estimator of $\mu_\ast(x)$; here, we are going to consider the empirical measure given by
\begin{equation}
\label{eqn:empirical-measure}
    \mu_n(x) = \frac{1}{n} \sum_{i=1}^n \delta_{X_i}(x),
\end{equation}
where $\delta_{X_i}(\cdot)$ denotes the Dirac measure at $X_i$.
The sample version of $HD^2$ is given by 
\begin{align*}
  \label{equ:hd-complete-sample}
  HD^2_n(\theta,\alpha) = & \int D_\alpha(x) d\mu_n(x) + \int (1-D_{\alpha,\theta} (z)) d\mu_g(z) \\
  & - 2\int D^{\frac{1}{2}}_\alpha(x)d\mu_n(x) \int(1-D_{\alpha,\theta}(z))^{\frac{1}{2}}d\mu_g(z)\\
  = & h_{1,n}(\alpha) + h_{2}(\theta,\alpha) -2\gamma_n(\theta,\alpha).
\end{align*}
where
\begin{equation}
	\label{eq:obj-gamma-sample}
	\gamma_n(\theta,\alpha) = \int D^{\frac{1}{2}}_\alpha(x) d\mu_n(x)\int (1-D_\alpha(G_\theta(z)))^{\frac{1}{2}}  d\mu_g(z).
\end{equation}

\begin{remark}
    In the definition above, we adopt the empirical measure $\mu_n$ as a natural plug-in estimator of $\mu_\ast$. However, this choice is not unique. In particular, one can also approximate $\mu_\ast$ by a smoothed estimator such as the kernel density estimator (KDE)
    \begin{equation}
    \label{eqn:kde-hn}
      d\mu_n^{\mathrm{KDE}}(x) = h_n(x)d\mu(x), \qquad  h_n(x) = \frac{1}{n c^d_{n}}\sum_{i=1}^n K\left(\frac{x-X_i}{c_{n}}\right)
    \end{equation}
    with kernel function $K$ and bandwidth parameter $c_n$. 
    Using KDEs, the empirical measure leads to a direct empirical-process formulation which can be used to establish consistency and asymptotics. We have explored the theoretical properties of this alternative formulation and provide a detailed discussion in Section S1 of the Supplementary Material.
\end{remark}
Let $(\theta_n, \alpha_n)$ be the solution of the empirical objective function, that is
\begin{equation}
\label{eqn:solution-n}
    (\theta_n,\alpha_n)=\arg\inf_{\theta\in\Theta} \sup_{\alpha\in\Lambda} \mathrm{HD}^2_n(\theta,\alpha).
\end{equation}
We have the following result.
\begin{theorem}
    If $(H_D)$ and $(H_G)$ hold, then 
    \begin{equation*}
    (\theta_n,\alpha_n) \stackrel{a.s.}{\longrightarrow}(\theta_\ast,\alpha_\ast) \mbox{ as } n \to \infty.
\end{equation*}
\end{theorem}
\begin{proof}
The proof is divided into two parts. We first start showing that $\mathrm{HD}^2_n(\theta,\alpha)$ converges uniformly, almost surely, to $\mathrm{HD}^2(\theta,\alpha)$, i.e.
\begin{equation*}
   \lim_{n\to\infty} \sup_{(\theta,\alpha) \in \Theta \times \Lambda}|\mathrm{HD}^2_n(\theta,\alpha) - \mathrm{HD}^2(\theta,\alpha)| = 0.
\end{equation*}
Note that, since $D_\alpha(x)$ is bounded, by the Strong Law of Large Numbers  
\begin{align*}
    |h_{1,n}(\alpha) - h_1(\alpha)| & = \left| \int D_{\alpha}(x) d\mu_n(x) -\int D_\alpha(x) d\mu_\ast(x)\right| \\
    & = \left|\int D_{\alpha}(x) (d\mu_n(x)  - d\mu_\ast(x)) \right| \to 0 \quad \mbox{ as } n \to \infty.
\end{align*}
Now consider $\gamma_{n}$. Notice that
\begin{align*}
    \left| {\gamma}_{n}(\theta,\alpha) - \gamma(\theta,\alpha)\right| \le &  \left| \int D_{\alpha}^{\frac{1}{2}}(x) (d\mu_n(x) - d\mu_\ast(x)) \int (1-D_{\alpha, \theta}(z))^{\frac{1}{2}}  d\mu_g(z) \right|\\
    & \le \left|\int D_{\alpha}^{\frac{1}{2}}(x) (d\mu_n(x) - d\mu_\ast(x))\right| \quad \forall \theta \in \Theta, \quad \forall \alpha \in \Lambda.
\end{align*}
Hence, in a similar way as above this term converges almost surely to zero as $n\to \infty$.

\noindent Now, recall that $\Theta \times \Lambda$ is compact, then we can extract a convergent subsequence $(\theta_{n_k}, \alpha_{n_k})$ from any sequence $(\theta_n, \alpha_n)$, i.e.
\begin{equation*}
    (\theta_{n_k},\alpha_{n_k}) \longrightarrow (\tilde{\theta}, \tilde{\alpha}) .
\end{equation*}
where $ (\tilde{\theta}, \tilde{\alpha})$ is the limiting value. By the continuity of $\mathrm{HD}^2$, we have that 
\begin{equation*}
    \lim_{n\to\infty} | \mathrm{HD}^2(\theta_{n_k},\alpha_{n_k}) - \mathrm{HD}^2(\tilde{\theta},\tilde{\alpha}) | = 0.
\end{equation*}
Note that
\begin{align*}
|\mathrm{HD}_{n_k}^2(\theta_{n_k}, \alpha_{n_k}) - \mathrm{HD}^2(\tilde{\theta}, \tilde{\alpha})|  \le & |\mathrm{HD}_{n_k}^2(\theta_{n_k}, \alpha_{n_k}) - \mathrm{HD}^2(\theta_{n_k}, \alpha_{n_k})|  \\
& + |\mathrm{HD}^2(\theta_{n_k}, \alpha_{n_k}) - \mathrm{HD}^2(\tilde{\theta}, \tilde{\alpha}) | \\
\le & \sup_{(\theta,\alpha) \in \Theta \times \Lambda} |\mathrm{HD}_{n_k}^2(\theta, \alpha) - \mathrm{HD}^2(\theta, \alpha)| \\
& + |\mathrm{HD}^2(\theta_{n_k}, \alpha_{n_k}) - \mathrm{HD}^2(\tilde{\theta}, \tilde{\alpha}) | .  
\end{align*}
and since we proved that both terms in the right hand side go to zero as $n \to \infty$, then 
\begin{equation*}
    \mathrm{HD}_{n_k}^2(\theta_{n_k},\alpha_{n_k}) \longrightarrow \mathrm{HD}^2(\tilde{\theta},\tilde{\alpha}) \qquad \mu-a.s..
\end{equation*}
Notice that $(\theta_{n_k},\alpha_{n_k})$ is the optimizer of $\mathrm{HD}^2_{n_k}(\theta,\alpha)$, then $(\tilde{\theta},\tilde{\alpha})$ is the optimizer of $\mathrm{HD}^2(\theta,\alpha)$. For the uniqueness, we have $(\tilde{\theta},\tilde{\alpha})\equiv(\theta_\ast,\alpha_\ast)$.
\end{proof}

\subsection{Joint Asymptotic Normality}

We now investigate the joint asymptotic normality of $(\theta_n,\alpha_n)$. The optimizer $(\theta_n,\alpha_n)$ is solution of the estimating equations given by 
$$\nabla \mathrm{HD}^2_n(\theta,\alpha) = 0$$ 
where $\nabla$ denotes the gradient operator with respect to $\theta$ and $\alpha$, i.e. 
$\nabla = \begin{pmatrix}
\nabla_\alpha \\
\nabla_\theta
\end{pmatrix}.$ 
By Taylor's series expansion of $\mathrm{HD}^2_n(\theta,\alpha)$ around $(\theta_\ast,\alpha_\ast)$, we get
\begin{equation}
\nabla \mathrm{HD}^2_n(\theta_n,\alpha_n) = \nabla \mathrm{HD}^2_n(\theta_\ast,\alpha_\ast) + \nabla^2 \mathrm{HD}_n^2(\theta_n^\ast,\alpha_n^\ast) [(\theta_n,\alpha_n) - (\theta_\ast,\alpha_\ast)]
\end{equation}
where $\nabla^2$ denotes the matrix of second derivatives, i.e.
$$\nabla^2 = \begin{pmatrix}
\nabla_\alpha^2 & \nabla_\theta \nabla_\alpha\\
\nabla_\alpha \nabla_\theta & \nabla_\theta^2
\end{pmatrix},$$
and $(\theta_n^\ast,\alpha_n^\ast) \in \mathcal{U}_n(\theta_\ast)\times\mathcal{V}_n(\alpha_\ast)$ with $\mathcal{U}_n(\theta_\ast) = \{ \theta| \theta = t \theta_\ast + (1-t) \theta_n\}$ and $\mathcal{V}_n(\alpha_\ast) = \{ \alpha| \alpha = t \alpha_\ast + (1-t) \alpha_n\}$. 
Note that $\nabla \mathrm{HD}^2_n(\theta_n,\alpha_n) = 0$, hence
\begin{equation}
    (\theta_n,\alpha_n) - (\theta_\ast,\alpha_\ast) = - [\nabla^2 \mathrm{HD}_n^2(\theta_n^\ast,\alpha_n^\ast)]^{-1} \nabla \mathrm{HD}^2_n(\theta_\ast,\alpha_\ast).
\end{equation}
We now introduce some lemmas that will be useful in determining the joint asymptotic distribution of $\sqrt{n}((\theta_n, \alpha_n) - (\theta_\ast, \alpha_\ast))$.

\begin{lemma}
\label{lemma:1}
    Consider $X_1, \ldots, X_n$ i.i.d. observations such that $X_i \sim p_\ast$ and let $\mu_n(x)$ be the empirical estimator given in equation (\ref{eqn:empirical-measure}). 
    Let $f$ be a $d$-dimensional function uniformly bounded. We have that
\begin{itemize}
\item[(i)]
  \begin{equation*}
      \int f(x) \big(d\mu_{n}(x)-d\mu_{\ast}(x)\big)  \stackrel{p}{\longrightarrow} 0 \qquad \mbox{ as } n \to \infty;
  \end{equation*}
  \item[(ii)] 
  \begin{equation*}
      \sqrt{n} \int f(x)\big(d\mu_{n}(x)-d\mu_{\ast}(x)\big) \stackrel{d}{\longrightarrow} N_d (0, \Sigma_f) \qquad \mbox{ as } n \to \infty,
  \end{equation*}
  where $\Sigma_f = \mathrm{Var}[f(X)]$.
\end{itemize}
\end{lemma}

Consider the following assumptions:
\begin{itemize}
    \item[($H_1$)] $(\alpha, x) \to D_\alpha(x)$ is of class $\mathcal{C}^2$, uniformly bounded with uniformly bounded differential of first and second order.
    \item[($H_2$)] $\forall z \in \mathbb{R}^{d^\prime}$, $\theta \to G_\theta(z)$ is of class $\mathcal{C}^2$, uniformly bounded with uniformly bounded differential.
    \item[($H_3$)] The Hessian matrix of the objective function, $\nabla^2 \mathrm{HD}^2(\theta, \alpha)$, is positive definite at the true parameter values $(\theta^*, \alpha^*)$. 
\end{itemize}

\begin{proposition}
\label{prop1}
    Assume that $(H_1)-(H_3)$ hold, then %, and the assumptions of the lemmas are satisfied. Then
    \begin{equation*}
        \sqrt{n} \nabla \mathrm{HD}_n^2(\theta_\ast, \alpha_\ast) \longrightarrow N(0, S),
    \end{equation*}
    where $S$ is a non-singular covariance matrix.
\end{proposition}
\begin{proof}

The derivative with respect to $\theta$ given in Appendix \ref{app:derivatives1} computed at $(\theta_\ast, \alpha_\ast)$ can be rewritten as 
\begin{align*}
    \sqrt{n}\nabla_\theta &\mathrm{HD}^2_n(\theta_\ast,\alpha_\ast) =  \\
    & - \sqrt{n}\int \nabla_\theta D_{\alpha_\ast,\theta_\ast} (z) d\mu_g(z) \\
    & + \sqrt{n}\int D_{\alpha_\ast}^{\frac{1}{2}}(x) (d\mu_n(x) - d\mu_\ast(x)) \int\frac{\nabla_\theta D_{\alpha_\ast,\theta_\ast}(z)}{(1-D_{\alpha_\ast,\theta_\ast}(z))^{1/2}}  d\mu_g(z) \\
    & + \sqrt{n}\int D_{\alpha_\ast}^{\frac{1}{2}}(x) d\mu_\ast(x) \int\frac{\nabla_\theta D_{\alpha_\ast,\theta_\ast}(z)}{(1-D_{\alpha_\ast,\theta_\ast}(z))^{1/2}} d\mu_g(z) \\
    =& T_{1} + T_{2n} + T_{3}.
\end{align*}
By Lemma \ref{lemma:1}, we have that
\begin{equation*}
    T_{2n}  \stackrel{d}{\longrightarrow} N(0, S_1),
\end{equation*}
with
\begin{equation*}
    \Delta_1(X) =
    \left(D_{\alpha_\ast}^{\frac{1}{2}}(X)\right) \E_{Z\sim g} \left[\frac{\nabla_\theta D_{\alpha_\ast,\theta_\ast}(Z)}{(1-D_{\alpha_\ast,\theta_\ast}(Z))^{1/2}}\right].
\end{equation*}
and 
\begin{align*}
    S_1 = & \mathrm{Var}(\Delta_1(X)) \\
    = & \E_{Z\sim g} \left[\frac{\nabla_\theta D_{\alpha_\ast,\theta_\ast}(Z)}{(1-D_{\alpha_\ast,\theta_\ast}(Z))^{1/2}}\right] \mathrm{Var}
    \left(D_{\alpha_\ast}^{\frac{1}{2}}(X)\right) \E_{Z\sim g} \left[\frac{\nabla_\theta D_{\alpha_\ast,\theta_\ast}(Z)}{(1-D_{\alpha_\ast,\theta_\ast}(Z))^{1/2}}\right]^\top.
\end{align*}
Notice that $T_{1} + T_{3} = 0$ since it corresponds to $\nabla_\theta \mathrm{HD}^2(\theta_\ast, \alpha_\ast) = 0$.

\noindent Let us now consider the derivative with respect to $\alpha$ given in Appendix \ref{app:derivatives1} computed at $(\theta_\ast, \alpha_\ast)$, that can be rewritten as  
\begin{align*}
    \sqrt{n}\nabla_\alpha &\mathrm{HD}^2_n(\theta_\ast,\alpha_\ast) =\\
    & + \sqrt{n}\int \nabla_\alpha D_{\alpha_\ast}(x) (d\mu_n(x) - d\mu_\ast(x)) \\
    & + \sqrt{n}\int \nabla_\alpha D_{\alpha_\ast}(x) d\mu_\ast(x) \\
    & - \sqrt{n}\int \nabla_\alpha D_{\alpha_\ast,\theta_\ast}(z)  d\mu_g(z) \\
    & - \sqrt{n}\int \frac{\nabla_\alpha D_{\alpha_\ast}(x)}{D_{\alpha_\ast}(x)^{1/2}} (d\mu_n(x) - d\mu_\ast(x)) \int(1-D_{\alpha_\ast,\theta_\ast}(z))^{\frac{1}{2}} d\mu_g(z) \\
    & - \sqrt{n}\int \frac{\nabla_\alpha D_{\alpha_\ast}(x)}{D_{\alpha_\ast}(x)^{1/2}} d\mu_\ast(x) \int(1-D_{\alpha_\ast,\theta_\ast}(z))^{\frac{1}{2}}  d\mu_g(z) \\
    & + \sqrt{n}\int D_{\alpha_\ast}^{\frac{1}{2}}(x) (d\mu_n(x) - d\mu_\ast(x))  \int\frac{\nabla_\alpha D_{\alpha_\ast,\theta_\ast}(z)}{(1-D_{\alpha_\ast,\theta_\ast}(z))^{1/2}} d\mu_g(z)\\
    & + \sqrt{n}\int D_{\alpha_\ast}^{\frac{1}{2}}(x) d\mu_\ast(x) \int\frac{\nabla_\alpha D_{\alpha_\ast,\theta_\ast}(z)}{(1-D_{\alpha_\ast,\theta_\ast}(z))^{1/2}} d\mu_g(z)\\
    =& J_{1,n} + J_{2} + J_{3} + J_{4,n} + J_{5} + J_{6,n} + J_{7}.
\end{align*}
Notice that $J_{2} + J_{3} + J_{5} + J_{7} = 0$ since it corresponds to $\nabla_\alpha \mathrm{HD}^2(\theta_\ast, \alpha_\ast) = 0$. The remaining terms can be rewritten as 
\begin{align*}
    J_{1,n} + J_{4,n} + J_{6,n} = \sqrt{n}\int \Delta_2(x) (d\mu_n(x) - d\mu_\ast(x))  
\end{align*}
where   
\begin{align*}
    \Delta_2(X) =& \nabla_\alpha D_{\alpha_\ast}(X) - \frac{\nabla_\alpha D_{\alpha_\ast}(X)}{D_{\alpha_\ast}(X)^{1/2}} \E_{Z\sim g} \left[(1-D_{\alpha_\ast,\theta_\ast}(Z))^{\frac{1}{2}} \right] \\
    & + D_{\alpha_\ast}^{\frac{1}{2}}(X) \E_{Z\sim g} \left[ \frac{\nabla_\alpha D_{\alpha_\ast,\theta_\ast}(Z)}{(1-D_{\alpha_\ast,\theta_\ast}(Z))^{1/2}} \right]
\end{align*}
By Lemma \ref{lemma:1}, we have that
\begin{equation*}
    J_{1,n} + J_{4,n} + J_{6,n}  \stackrel{d}{\longrightarrow} N(0, S_2) \quad\mbox{ with } \quad S_2 = \mathrm{Var}(\Delta_2(X)).
\end{equation*}
Then by the central limit theorem and the continuous mapping theorem, we have that
\begin{equation*}
    (T_{2n}; J_{1n} + J_{4n} + J_{6n}) \longrightarrow N(0, S) 
\end{equation*}
where $S = \begin{bmatrix}
S_1 & S_{12} \\
S_{12}^\top & S_2
\end{bmatrix}$ 
where $S_{12} = \mathrm{Cov}(\Delta_1(X), \Delta_2(X))$.
\end{proof}

\begin{proposition}
\label{prop2}
    Assume that $(H_1)-(H_3)$ are satisfied. Then 
    \begin{equation*}
        \lim_{n\to \infty} \nabla^2 \mathrm{HD}_n^2(\theta_n^\ast, \alpha_n^\ast) = \nabla^2 \mathrm{HD}^2(\theta_\ast, \alpha_\ast) 
    \end{equation*}
\end{proposition}
\begin{proof}

The idea is to show the convergence element-wise, considering the second derivatives separately, which are reported in Appendix \ref{app:derivatives2}. 
Consider the second derivative with respect to $\theta$. Notice that, using simple operations
\begin{align*}
    A_2 =& \int D_{\alpha}^{\frac{1}{2}}(x) (d\mu_n(x) - d\mu_\ast(x)) \int \frac{\nabla_{\theta \theta^\top} D_{\alpha, \theta}(z)}{(1 - D_{\alpha, \theta}(z))^{\frac{1}{2}}} d\mu_g(z)\\
    & + \int D_{\alpha}^{\frac{1}{2}}(x) d\mu_\ast(x) \int \frac{\nabla_{\theta \theta^\top} D_{\alpha, \theta}(z)}{(1 - D_{\alpha, \theta}(z))^{\frac{1}{2}}} d\mu_g(z).
\end{align*}
Note that, by $(H_2)$, $D_\alpha^{\frac{1}{2}}(x)$ and $\nabla_{\theta \theta^\top} D_{\alpha,\theta}(z)$ are bounded and continuous around $\theta_\ast$ for all $z \in \mathbb{R}^{d^\prime}$. Then, the first term converges to zero as $n \to \infty$ by the point $(i)$ of the Lemma \ref{lemma:1}, while the second term does not depend on $n$. Similarly
\begin{align*}
    A_3 =& \frac{1}{2} \int D_{\alpha}^{\frac{1}{2}}(x) (d\mu_n(x) - d\mu_\ast(x)) \int \frac{\nabla_\theta D_{\alpha, \theta}(z) \nabla_\theta^\top D_{\alpha, \theta}(z)}{(1 - D_{\alpha, \theta}(z))^{\frac{3}{2}}} d\mu_g(z)\\
    & \frac{1}{2} \int D_{\alpha}^{\frac{1}{2}}(x) d\mu_\ast(x) \int \frac{\nabla_\theta D_{\alpha, \theta}(z) \nabla_\theta^\top D_{\alpha, \theta}(z)}{(1 - D_{\alpha, \theta}(z))^{\frac{3}{2}}} d\mu_g(z). 
\end{align*}
By Lemma \ref{lemma:1} the first term converges to zero. 
Analogous results can be obtained for the other terms as well as for the other derivatives. See full details in Section S2 of the Supplementary Material. 
\end{proof}
To conclude, we state the asymptotic normality of $(\theta_n, \alpha_n)$. 
\begin{theorem}
\label{th:asym-norm-complete}
  Assume that assumptions $(H_D), (H_G), (H_1)-(H_3)$ hold. Let $(\theta_n,\alpha_n)$ be the sequence of estimators defined in \ref{eqn:solution-n} and let $(\theta^\ast,\alpha^\ast)$ denote the unique minimax solution of equation \ref{eqn:solution-star}. Then, as $n \to \infty$
  \begin{equation*}
      \sqrt{n} ((\theta_n,\alpha_n) -(\theta_\ast,\alpha_\ast)) \stackrel{d}{\rightarrow} N(0, \Sigma)
  \end{equation*}
  where
  \begin{equation*}
    \Sigma =  J^{-1} S  (J^{-1})^{\top}, 
    \qquad
    J = \nabla^2 \mathrm{HD}^2(\theta_\ast,\alpha_\ast),
  \end{equation*}
  and $S$ is the covariance matrix in Proposition~\ref{prop1}.
\end{theorem}
\begin{proof}
Combining the results of Proposition \ref{prop1} and Proposition \ref{prop2} the theorem follows.
\end{proof}

\begin{remark}
An alternative way to prove these asymptotic properties for the proposed Hellinger-based losses is to consider the profiled version of our estimator that focuses directly on the generator parameter. Specifically, we consider the profiled objectives
$$
S_n(\theta) = \sup_{\alpha\in\Lambda} \HD_n^2(\theta,\alpha),
\qquad 
S(\theta) = \sup_{\alpha\in\Lambda} \HD^2(\theta,\alpha),
$$
and study the profiled estimator $\hat\theta_n\in\arg\min_{\theta\in\Theta} S_n(\theta)$. Under the same regularity conditions $(H_D)$, $(H_G)$ and $(H_1)-(H_3)$, we show that $S_n$ converges to $S$ uniformly on $\Theta$, which yields almost sure consistency of $\hat\theta_n$ for the unique minimizer $\theta^\ast$ of $S$. Moreover, by combining envelope and implicit-function arguments, we derive a central limit theorem for the profiled estimator
$$
\sqrt{n}\,(\hat\theta_n-\theta^\ast)\ \longrightarrow\ \mathcal N(0,\Sigma_{\theta}),
$$
where $\Sigma_{\theta}$ coincides with the $\theta$–marginal of the joint asymptotic covariance matrix in Theorem~\ref{th:asym-norm-complete}. The same conclusions hold for the approximated Hellinger objective. Full statements and proofs are reported in Section S4 of the Supplementary Material. 
\end{remark}

\section{Influence Function}
\label{sec:IF}

In this section, we compute the influence functions (IFs) associated with the classical GAN loss and the Hellinger-type loss proposed in this work. 
The IF offers valuable information on the robustness properties of adversarial training procedures, as it characterizes the local sensitivity of the adversarial game to infinitesimal contamination at a given point. Intuitively, it quantifies how a small perturbation in the data affects the resulting parameter estimates. 

Let $p_\varepsilon (x) = (1 - \varepsilon) q_{\theta_0}(x)  + \varepsilon h(x)$ denote the contaminated density at $x$, with the contaminating distribution $h(x)$ and let $\thetaz$ denote the parameter values for which the model density $q_{\thetaz}$ coincides with the true distribution $p_\ast$. 
Given the underlying random vector $Z \sim g$, we have $X \sim p_\varepsilon(x)$ the contaminated random vector that
generates the data.
For $\varepsilon = 0$ we have $X = G_{\thetaz}(Z)$. %Note that $X$ and $X_\varepsilon$ coincide only in $\varepsilon = 0$.
For all $\varepsilon\in[0,1)$, we define the Hellinger loss function under contamination given as
\begin{align*}
  \HD^2_\varepsilon(\theta, \alpha) & = \int D_\alpha(x) p_\varepsilon(x) \dx 
 + \int (1 - D_\alpha(x)) q_\theta(x) \dx 
   - 2 \gamma_\varepsilon(\theta, \alpha)
\end{align*}
where
\begin{equation*}
\gamma_\varepsilon(\theta, \alpha) = \left( \int D_\alpha(x)^{1/2} p_\varepsilon(x) \dx \right) \left( \int (1 - D_\alpha(x))^{1/2} q_\theta(x) \dx \right) = C_1C_2\ .
\end{equation*}
For each $\varepsilon \in [0, 1)$, we define 
\begin{equation*}
(\thetae, \alphae) = \arg\inf_\theta\sup_\alpha \HD^2_\varepsilon(\theta, \alpha).
\end{equation*}
Notice that for $\varepsilon = 0$, $\HD^2_0(\theta, \alpha)$ denotes the uncontaminated objective function, and $(\theta_\ast, \alpha_\ast)$ is the corresponding solution.
Observe that $p_\varepsilon(x)$ denotes the contaminated distribution, while $q_{\theta_\varepsilon}(x)$ is the model density evaluated at the parameter estimated under contamination.

Then, we define the influence functions for the Hellinger loss as
$$(IF(\theta), IF(\alpha)) =  \left.\frac{\partial}{\partial\varepsilon}(\thetae, \alphae)\right|_{\varepsilon = 0}.$$
The optimizer $(\thetae, \alphae)$ is solution of $\nabla \HD^2_\varepsilon(\thetae, \alphae) = 0$, for a fixed $\varepsilon \in [0,1)$. Hence, in order to compute the influence function we can consider, for all $\varepsilon \in [0,1)$
\begin{equation*}
    \frac{\partial}{\partial \varepsilon} \nabla \HD^2_\varepsilon(\theta_\varepsilon, \alpha_\varepsilon) = 0\ .
\end{equation*}
We have
\begin{align*}
\nabla_\alpha \HD^2_\varepsilon(\alpha, \theta) & = \int \nabla_\alpha D_\alpha(x) p_\varepsilon(x) \dx  - \int \nabla_\alpha D_\alpha(x) q_\theta(x) \dx  - \nabla_\alpha \gamma_\varepsilon(\alpha, \theta) \\
 & = A_\alpha + B_\alpha - 2(C_{1\alpha} C_2 + C_1 C_{2\alpha})
\end{align*}
and
\begin{align*}
\nabla_\theta \HD^2_\varepsilon(\alpha, \theta) & = \int (1 - D_\alpha(x)) s_\theta(x) q_\theta(x) \dx - \nabla_\theta \gamma_\varepsilon(\alpha, \theta) \\
& = B_\theta - 2 C_1 C_{2\theta}\ ,
\end{align*}
where $C_{1\alpha} = \nabla_\alpha C_1$, $C_{2\alpha} = \nabla_\alpha C_2$, $C_{2\theta} = \nabla_\theta C_2$, and $s_\theta(x) = \nabla_\theta \log q_\theta(x)$ is the usual score function. 
We need to compute the derivatives with respect to $\varepsilon$ of these terms and evaluate the expressions at $\varepsilon = 0$.
For the first term
\begin{align*}
\frac{\partial}{\partial \varepsilon} A_\alpha & = \int \frac{\partial}{\partial \varepsilon} \nabla_\alpha D_\alphae(x) p_\varepsilon(x) \dx + \int \nabla_\alpha D_\alphae(x) \frac{\partial}{\partial \varepsilon} p_\varepsilon(x) \dx \\
& = \int \nabla_{\alpha\alpha^\top} D_\alphae(x) p_\varepsilon(x) \dx \frac{\partial}{\partial \varepsilon} \alpha_\varepsilon + \int \nabla_\alpha D_\alphae(x) (h(x) - q_\thetaz(x)) \dx \\
& \text{with $\varepsilon = 0$ we obtain} \\
& = \int \nabla_{\alpha\alpha^\top} D_\alphaz(x) q_\thetaz(x) \dx \IF(\alpha) + \int \nabla_\alpha D_\alphaz(x) (h(x) - q_\thetaz(x)) \dx \\
& = A_{1\alpha} \IF(\alpha) + A_{2\alpha}
\end{align*}
Following similar steps, the derivatives computed at $\varepsilon = 0$ are given as
\begin{align*}
% B-alpha
\frac{\partial}{\partial \varepsilon} B_\alpha  = & - \int \nabla_{\alpha\alpha^\top} D_\alphaz(x) q_\thetaz(x) \dx \IF(\alpha) - \int \nabla_\alpha D_\alphaz(x) s_\thetaz^\top(x) q_\thetaz(x) \dx \IF(\theta) \\
 =&  B_{1\alpha} \IF(\alpha) + B_{2\alpha} \IF(\theta)\ ; \\
% B-theta
\frac{\partial}{\partial \varepsilon} B_\theta  = & - \int \nabla_{\alpha} D_\alphaz(x) s_\thetaz(x) q_\thetaz(x) \dx \IF(\alpha) + \int (1 - D_\alphaz(x)) \frac{\nabla_{\theta\theta^\top} q_\thetaz(x)}{q_\thetaz(x)} q_\thetaz(x) \dx \IF(\theta) \\
 = & \ B_{1\theta} \IF(\alpha)  + B_{2\theta} \IF(\theta) \ ; \\
%C-1alpha
\frac{\partial}{\partial \varepsilon} C_{1\alpha} = & \int D_\alphaz(x)^{-1/2} \left(\nabla_{\alpha,\alpha^\top} D_\alphaz(x) - \frac{1}{2}  D_\alphaz(x)^{-1} \nabla_\alpha D_{\alphaz}(x) \nabla_\alpha D_{\alphaz}(x)^\top \right) q_\thetaz(x) \dx IF(\alpha) \\
& + \int D_\alphaz(x)^{-1/2} \nabla_\alpha D_\alphaz(x) (h(x) - q_\thetaz(x)) \dx \\
= & \ C_{1\alpha a} IF(\alpha) + C_{1\alpha b}\ ; \\
%C-2
\frac{\partial}{\partial \varepsilon} C_2  = & - \frac{1}{2} \int (1 - D_\alphaz(x))^{-1/2} \nabla_\alpha D_\alphaz(x)^\top q_\thetaz(x) \dx \IF(\alpha) \\ 
& + \int (1 - D_\alphaz(x))^{1/2} s_\thetaz(x) q_\thetaz(x) \dx \IF(\theta) \\
= & \ C_{2a} IF(\alpha) + C_{2b}IF(\theta) \ ;\\
%C-1
\frac{\partial}{\partial \varepsilon} C_1  = & \frac{1}{2} \int D_\alphaz^{-1/2}(x) \nabla_\alpha D_\alphaz(x)^\top q_{\thetaz}(x) \dx IF(\alpha) + \int D_\alphaz(x)^{1/2} (h(x) - q_\thetaz(x)) \dx\\
= & \ C_{1a} IF(\alpha) + C_{1b} ;\\
%C-2alpha
\frac{\partial}{\partial \varepsilon} C_{2\alpha}  = &- \frac{1}{2} \int (1 - D_\alphaz(x))^{-3/2} \nabla_\alpha D_\alphaz(x) \nabla_\alpha D_\alphaz(x)^\top q_\thetaz(x) \dx IF(\alpha) \\
& + \int (1 - D_\alphaz(x))^{-1/2} \nabla_{\alpha,\alpha^\top} D_\alphaz(x) q_\thetaz(x) \dx IF(\alpha) \\
& + \int (1 - D_\alphaz(x))^{-1/2} \nabla_\alpha D_\alphaz(x) s_\thetaz(x)^\top q_\thetaz(x) \dx IF(\theta) \\
= & \ C_{2\alpha a} IF(\alpha) + C_{2\alpha b} IF(\alpha) + C_{2\alpha c} IF(\theta); \\
%C-2theta
\frac{\partial}{\partial \varepsilon} C_{2\theta}  = & \int (1 - D_\alphaz(x))^{1/2} \frac{\nabla_{\theta \theta^\top} q_\thetaz(x)}{q_\thetaz(x)}  q_\thetaz(x) \dx IF(\theta) \\
& - \frac{1}{2} \int \frac{\nabla_{\alpha} D_\alphaz(x)}{(1 - D_\alphaz(x))^{\frac{1}{2}}}  q_\thetaz(x) \dx IF(\alpha) \\
= & \ C_{2\theta a} IF(\theta) + C_{2\theta b} IF(\alpha) \ .
\end{align*}
Combining all the terms together, we have the following two equations
\begin{align*}
    0 = & \left[A_{2\alpha} - 2 C_{1\alpha b} C_2 - 2 C_{2\alpha}C_{1b}\right] \\
    & + \left[A_{1\alpha} + B_{1\alpha} - 2 C_{1\alpha a}C_2 - 2 C_{1\alpha} C_{2a} - 2 C_{2\alpha} C_{1a} -2  C_1 C_{2\alpha a} - 2  C_1 C_{2\alpha b} \right] IF(\alpha) \\
    & + \left[ B_{2\alpha} -2 C_{1\alpha} C_{2b} - 2 C_1 C_{2\alpha c}\right]\IF(\theta) \\
     = & I_0 + I_\alpha \IF(\alpha) + I_\theta \IF(\theta) ,
\end{align*}
and
\begin{align*}
    0 = & -2 C_{2\theta}C_{1b} + \left[B_{1\theta} -2 C_{2\theta }C_{1a} -2 C_{1} C_{2\theta b}  \right] IF(\alpha) + \left[ B_{2\theta} -2 C_{1} C_{2\theta a}\right]\IF(\theta) \\
     = & K_0 + K_\alpha \IF(\alpha) + K_\theta \IF(\theta).
\end{align*}
Solving the system we have
\begin{align*}
\IF(\alpha) & = (I_\alpha - I_\theta K_\theta^{-1} K_\alpha)^{-1} (I_\theta K_\theta^{-1}K_0 - I_0) \\
\IF(\theta) & = - K_\theta^{-1} (K_0 + K_\alpha \IF(\alpha)) =  - K_\theta^{-1} (K_0 + K_\alpha (I_\alpha - I_\theta K_\theta^{-1} K_\alpha)^{-1} (I_\theta K_\theta^{-1}K_0 - I_0)) \ .
\end{align*}
The computation of the influence function for the standard loss function is reported in Section S3 of the Supplementary Material. %In the next section, we also explore the influence functions numerically. 
Notice that the influence functions are calculated at the true parameter values $(\theta_\ast, \alpha_\ast)$. In the numerical experiments, we can set the true generator parameters $\theta_\ast$, however, it is not straightforward for the discriminator parameters. In fact, we do not know the true values $\alpha_\ast$ and we can only consider the estimated parameter $\hat\alpha$ instead. 

\section{Numerical Experiments}
\label{sec:simulation}

We conducted numerical experiments to evaluate the empirical performance of the proposed GAN framework with Hellinger-type loss functions. 
We considered data generated from the normal distribution $\mathcal{N}(\mu_0, \sigma_0)$ with $\mu_0=10$ and $\sigma_0=1.5$. The generator is a parametric normal model with unknown mean and variance, that is $G_\theta(z) = \sigma z + \mu$ with $\theta = (\mu, \sigma)$ and $Z \sim N(0, 1)$. The discriminator is implemented as a feedforward neural network with one hidden layer of five nodes (16 parameters in total). 
In the data generating setting, we considered the sample size $n=100000$, divided into batches of $n_B = 1000$ observations. 
This simplified setting is chosen since it allows for an explicit comparison of the parameter estimation quality and the influence of the loss function on the learning process.
We also investigate the performance of the proposed loss functions in the case of contamination. Specifically, a proportion $\varepsilon$ of data points is sampled from $\mathcal{N}(0, 1)$, at the percentage of contamination of $\varepsilon = 0, 1, 5, 10, 20$.
For each setting and contamination level, we perform $100$ independent replications, and each GAN is trained for 400 epochs using the Adam optimizer with learning rates 0.001, for both generator and discriminator. 

We track the following evaluation metrics:
\begin{itemize}
    \item Mean Squared Error (MSE) between the estimated generator parameters and the true values;
    \item The Root Mean Square Error (RMSE) for the generator parameters $\theta_\ast = (\mu_\ast, \sigma_\ast)$
    $$\mathrm{RMSEC}(\hat{\theta}) = \sqrt{\frac{1}{2} \left( \left(\hat{\mu} - \mu_\ast \right)^2 + \left(\hat{\sigma} - \sigma_\ast \right)^2\right)};$$
\end{itemize}
We compare the proposed approximated Hellinger loss in equation (\ref{eq:hat-gamma}) and the complete Hellinger-type loss in equation (\ref{eq:hd-complete}), considering the KDE as sample-based estimator with bandwidth parameter $h = 0.001, 0.01, 0.5$, with the standard GAN loss given in equation (\ref{eq:classical_loss}) and the WGAN. Our goal is to assess the convergence behavior and fidelity of the generated samples. 

\subsection{Results}
\label{subsec:results}

First, we examine the accuracy of parameter estimation for each GAN variant under varying contamination levels. Tables \ref{tab:best_mse_mu} and \ref{tab:best_mse_sigma} report the best median (standard deviation) MSE achieved for the generator’s parameters $\mu$ and $\sigma$, respectively, for each method. Similarly, Table \ref{tab:best_mse_combined} shows the best combined RMSE for $(\mu,\sigma)$ at the epoch with the lowest error. Here, the best epoch refers to the training epoch (out of 400) where the RMSE of the generator parameters was minimal; the corresponding MSE/RMSE values from that epoch are then averaged over replications. With this selection, we compare methods based on their best observed performance during training.

\begin{table}[htbp]
\def\arraystretch{1.3}
\caption{Median (standard deviation) of best MSE$(\hat{\mu})$ ($\times 100$) across replications for the considered methods and percentage of contamination $\varepsilon=0, 1, 5, 10, 20$.}
\label{tab:best_mse_mu}
\begin{adjustbox}{width=\textwidth}
\begin{tabular}{llllll}
\hline
 & $\varepsilon$=0\% & $\varepsilon$=1\% & $\varepsilon$=5\% & $\varepsilon$=10\% & $\varepsilon$=20\% \\
\hline
GAN & \textbf{0.001} (2.97) & \textbf{0.001} (1.44) & 0.035 (24.11) & 0.019 (82.30) & 141.103 (101.29) \\
WGAN & 18.567 (13.24) & 20.613 (17.38) & 3.359 (21.12) & 12.819 (8.57) & 26.588 (29.01) \\
Approx. HD & 0.002 (0.43) & \textbf{0.001} (0.30) & \textbf{0.002} (2.79) & \textbf{0.006} (15.53) & 93.003 (133.67) \\
HD ($c_n$=0.0001) & 0.042 (1.59) & 0.063 (689.19) & 0.052 (18.19) & 0.064 (86.67) & \textbf{0.203} (117.61) \\
HD ($c_n$=0.01) & 0.041 (1076.94) & 0.097 (130.29) & 0.223 (28.29) & 1.697 (47.63) & 70.500 (80.88) \\
HD ($c_n$=0.5) & 0.211 (1399.96) & 0.299 (0.53) & 0.221 (272.86) & 0.440 (0.73) & 29.548 (129.62) \\
\hline
\end{tabular}
\end{adjustbox}
\end{table}
\begin{table}[htbp]
\def\arraystretch{1.3}
\caption{Median (standard deviation) of best MSE$(\hat{\sigma})$ ($\times 100$) across replications for the considered methods and percentage of contamination $\varepsilon=0, 1, 5, 10, 20$.}
\label{tab:best_mse_sigma}
\begin{adjustbox}{width=\textwidth}
\begin{tabular}{llllll}
\hline
 & $\varepsilon$=0\% & $\varepsilon$=1\% & $\varepsilon$=5\% & $\varepsilon$=10\% & $\varepsilon$=20\% \\
\hline
GAN & \textbf{0.001} (74.14) & \textbf{0.001} (31.64) & 0.007 (340.75) & 0.010 (486.29) & 417.730 (706.51) \\
WGAN & 2.945 (2.19) & 5.052 (3.81) & 6.567 (8.78) & 15.325 (11.79) & 56.548 (64.32) \\
Approx. HD & 0.002 (19.06) & 0.002 (0.05) & \textbf{0.003} (58.65) & \textbf{0.007} (189.64) & 185.900 (612.16) \\
HD ($c_n$=0.0001) & 0.026 (0.93) & 0.058 (22.28) & 0.057 (105.66) & 0.057 (214.67) & \textbf{0.226} (473.39) \\
HD ($c_n$=0.01) & 0.101 (28.65) & 0.094 (16.21) & 0.081 (39.86) & 0.362 (97.98) & 42.283 (384.50) \\
HD ($c_n$=0.5) & 0.342 (30.38) & 0.305 (0.96) & 0.166 (233.43) & 0.085 (77.06) & 11.991 (319.05) \\
\hline
\end{tabular}
\end{adjustbox}
\end{table}
In the clean data setting, most of the methods perform well in terms of median error. The standard GAN and the proposed approximate Hellinger loss show extremely low MSE for $\mu$ and $\sigma$. The Hellinger GAN using a KDE-based loss with a very small bandwidth ($h=0.0001$) similarly attains a median error close to zero for both parameters. In contrast, the Wasserstein GAN shows a higher median error in the uncontaminated case, as well as the KDE-based Hellinger with a larger bandwidth. 
\begin{table}[htbp]
\def\arraystretch{1.3}
\caption{Median (standard deviation) of best RMSE$(\hat{\mu}, \hat{\sigma})$ ($\times 100$) across replications for the considered methods and percentage of contamination $\varepsilon=0, 1, 5, 10, 20$.}
\label{tab:best_mse_combined}
\begin{adjustbox}{width=\textwidth}
\begin{tabular}{llllll}
\hline
 & $\varepsilon$=0\% & $\varepsilon$=1\% & $\varepsilon$=5\% & $\varepsilon$=10\% & $\varepsilon$=20\% \\
\hline
GAN & \textbf{0.328} (36.68) & \textbf{0.300} (15.44) & 1.625 (85.08) & 1.232 (106.08) & 167.839 (135.95) \\
WGAN & 32.896 (10.13) & 35.885 (11.80) & 25.682 (18.53) & 39.399 (13.71) & 68.365 (18.82) \\
Approx. HD & 0.514 (9.94) & 0.410 (1.34) & \textbf{0.595} (17.65) & \textbf{0.926} (45.16) & 118.810 (118.35) \\
HD ($c_n$=0.0001) & 2.837 (4.66) & 2.924 (59.71) & 2.669 (37.20) & 2.620 (68.38) & \textbf{5.381} (102.47) \\
HD ($c_n$=0.01) & 3.026 (87.58) & 3.688 (27.93) & 4.084 (20.53) & 9.874 (32.16) & 103.390 (76.43) \\
HD ($c_n$=0.5) & 6.232 (109.08) & 6.444 (3.71) & 5.090 (52.67) & 5.680 (22.73) & 59.725 (72.82) \\
\hline
\end{tabular}
\end{adjustbox}
\end{table}
As the contamination level increases, the classical GAN loss becomes sensitive to even small fractions of outliers. By $\varepsilon=5\%$ and $10\%$, the standard GAN’s error even if shows low median MSE, begins to fluctuate substantially across runs, considering the large standard deviations, and at $\varepsilon=20\%$ its performance deteriorates drastically. The approximate Hellinger loss remains quite accurate up to moderate contamination but then degrades under higher contamination, showing a performance slightly better than the classical GAN. 
The WGAN shows a much more gradual increase in error as the contamination grows, with modest variability across runs suggesting consistent behavior even when outliers are present. Overall, while it sacrifices some absolute accuracy, WGAN has lower variability between replications.
The Hellinger GAN with KDE loss shows robustness that depends on the choice of bandwidth $c_n$. A very small bandwidth, $c_n=0.0001$, shows the lowest errors, even with $20\%$ contamination. 
Additionally, in terms of the combined RMSE, the approximate Hellinger model achieves a median RMSE which is slightly better than standard GAN, while the Hellinger GAN with KDE and $c_n=0.0001$ outperforms all the methods for high percentage of contamination. 
Tables S1--S3 in Section S5 of the Supplementary Material report the median MSE and RMSE of the generator parameters at the final epoch. 

\begin{figure}[htbp]
    \centering
    \includegraphics[width=\textwidth]{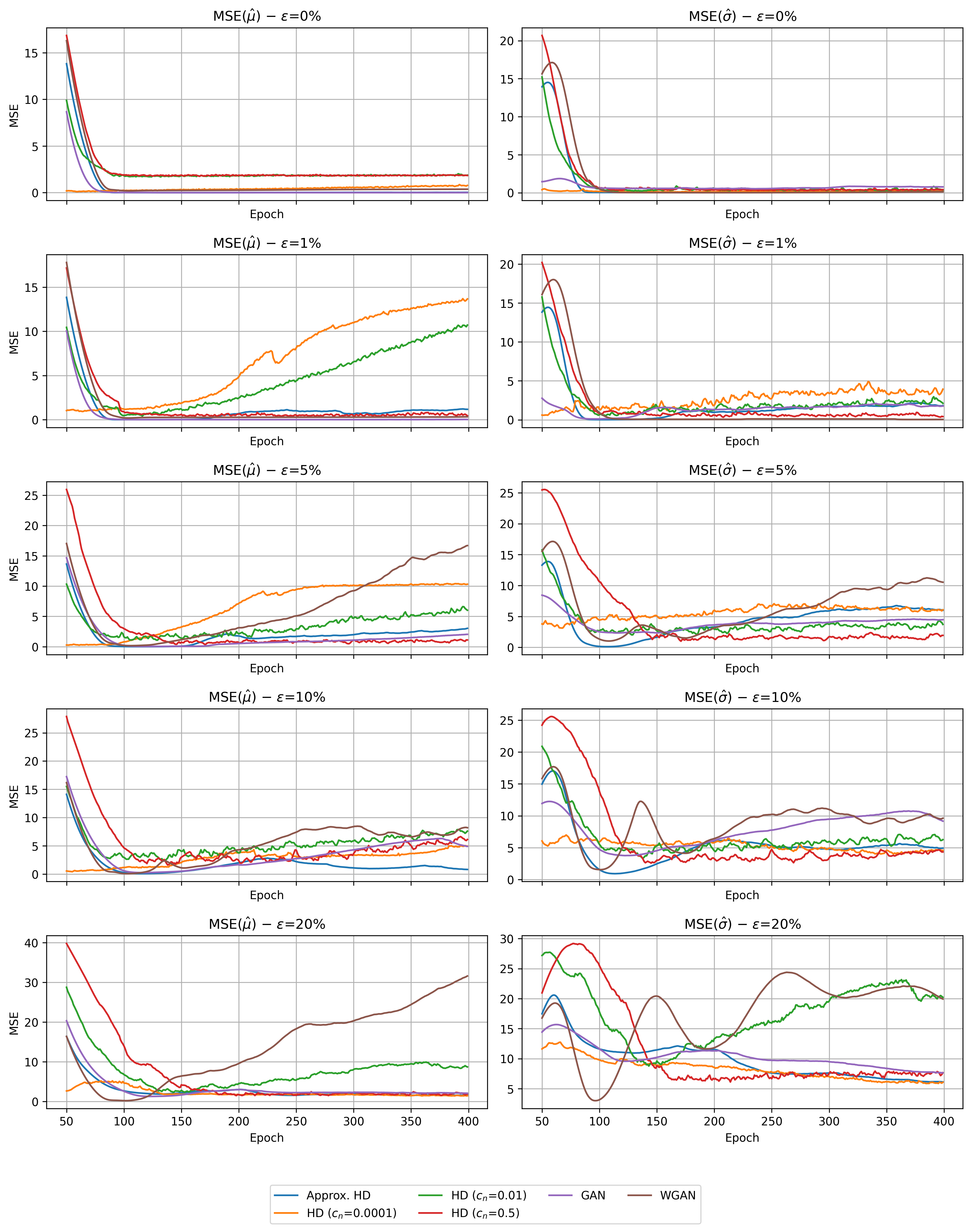}
    \caption{Mean Squared Error (MSE) per epoch for the different contamination percentage $\varepsilon=0, 1, 5, 10, 20$ comparing the different methods.}
    \label{fig:mse-subplots}
\end{figure}
Figure \ref{fig:mse-subplots} displays the evolution of the mean squared error (MSE) for the generator parameters $\mu$ (left column) and $\sigma$ (right column) over the training epochs, for the percentage of contamination $\varepsilon= 0, 5, 10, 20$. This visualization provides insight into the training dynamics of the GAN losses considered. Hellinger-type losses show robustness benefits over standard GAN and WGAN losses, both in terms of final accuracy and training stability.
In the uncontaminated case, most methods converge rapidly. For a higher level of contamination, the standard GAN shows larger errors for an increasing epoch, as well as WGAN, especially for $\sigma$. In contrast, the Approximate HD loss remains stable, with consistently low MSE throughout training even under 10\% contamination. The performance of KDE-based HD loss remains solid when the bandwidth is very small ($c_n$ = 0.0001). 

\section{Fashion MNIST dataset}
\label{sec:fashionMNIST}

Finally, we illustrate the proposed Hellinger losses on the higher dimensional Fashion MNIST image dataset. 
The training dataset is composed by 60,000 samples of $28\times28$ gray-scale images of clothing items from ten classes (e.g.\ T-shirt, trouser, coat, bag, boot).
We keep the original resolution and rescale pixel intensities to lie in $[0,1]$.  
The generator $G_\theta$ consists of three transpose–convolutional layers with batch normalization and ReLU activations that maps a latent vector $Z \sim \mathcal{N}(0, I_{100})$ to a $28\times 28$ image, while the discriminator $D_\alpha$ is a two–layer convolutional network with spectral normalization that maps an image to a scalar in $[0,1]$, using a sigmoid activation function.
We use Adam optimizer with learning rates of 0.0001 and update weight $\beta=0.5$.  We use a batch size of 128, sampled from the training dataset without replacement, and train the standard GAN, the approximated Hellinger GAN and the WGAN for $1000$ epochs. All models share the same network architectures, optimizer and learning-rates, while only the loss function is changed.

\begin{figure}[htb]
    \centering
    a) \includegraphics[width=.4\textwidth]{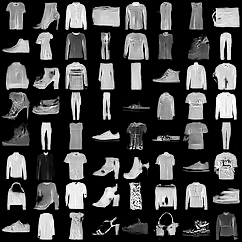}
    b)\includegraphics[width=.4\textwidth]{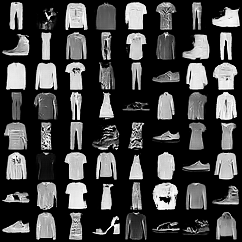}
    c)\includegraphics[width=.4\textwidth]{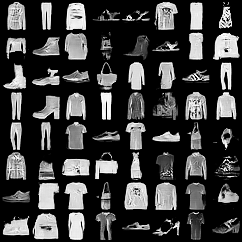}
    d)
    \includegraphics[width=.4\textwidth]{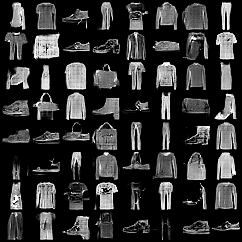}
    \caption{Fashion MNIST samples generated by the standard GAN (a), the approximated Hellinger GAN (b), the Hellinger GAN with KDE ($c_n=0.0001$) (c) and WGAN (d) after 1000 epochs.}
    \label{fig:fmnist_gan}
\end{figure}

Figure~\ref{fig:fmnist_gan} reports $8\times 8$ grids of samples generated after $1000$ epochs by the considered losses.  
The standard GAN produces visually plausible items with clear silhouettes. Similarly, the Hellinger-based losses yield images of comparable visual quality while preserving a high degree of diversity across different classes and styles.
By contrast, the WGAN configuration produces noticeably noisier samples. Many images show strong grid–like artifacts and saturated regions, and only a subset of silhouettes are clearly recognizable. This suggests that, with the present architecture and training setup, the Wasserstein objective is harder to optimize and yields a lower visual quality.

\section{Conclusion}
\label{sec:conclusion}

This work contributes to the statistical analysis of generative adversarial networks (GANs) by proposing and rigorously investigating a class of Hellinger-type loss functions within the GAN framework, motivated by the properties of symmetry, boundedness, and a natural connection to classical robust statistics.
We define an adversarial objectives that operates on the discriminator output and we study the resulting estimator within a parametric M-estimation framework. 
Under mild regularity assumptions, we develop a comprehensive asymptotic theory for the joint estimation of generator and discriminator parameters $(\hat\theta_n,\hat\alpha_n)$ under the proposed Hellinger-type loss. Our results establish the existence, uniqueness, consistency, and asymptotic normality of the estimators under mild regularity assumptions. 

We present controlled simulation experiments that highlight the advantages of the Hellinger-type GAN loss. In general, our simulation results demonstrate that in Gaussian settings the choice of loss function has an effect on training and robustness. The standard GAN is non-robust to even modest contamination, while the Wasserstein GAN is more resilient, maintaining bounded errors even as $\varepsilon$ increases. The proposed Hellinger-type losses achieve competitive accuracy in the uncontaminated case and maintain substantially lower and more stable mean squared errors under contamination, especially for the complete Hellinger loss with a properly calibrated bandwidth.  However, this depends on the choice of the bandwidth parameter $c_n$. The Fashion MNIST experiment, although not aimed at large-scale image generation, shows that the Hellinger-based losses can produce samples of comparable visual quality to the standard GANs, without sacrificing the robustness properties highlighted in the low-dimensional study.

This study highlights, among other aspects, that the choice of divergence in adversarial training affects the statistical properties of the resulting estimators. 
In general, distance-type losses can offer both practical and theoretical benefits in adversarial learning.
From a broader perspective, this work wants to highlight the importance of integrating statistical principles into the design of generative models. While recent advances have focused heavily on architectural innovation and empirical benchmarks, our findings suggest that studying the asymptotic properties of GAN estimators under various loss functions may lead to more reliable, interpretable, and robust data generation techniques.

\begin{appendix}

\section{}

\subsection{Derivatives of First order}
\label{app:derivatives1}

Here, we report the the first derivatives of the Hellinger loss, which are used in the proof of Proposition \ref{prop1}. Considering the following calculations
%\begin{itemize}
    %\item 
    $$\nabla(1 - D_{\alpha,\theta}(z)) = - \nabla D_{\alpha,\theta}(z),$$
    %\item 
    %$$\nabla_\theta(1 - D_{\alpha,\theta}(z)) = - \nabla_\theta D_{\alpha,\theta}(z)$$
    %\item 
    $$\nabla_\alpha(D_{\alpha,\theta}^{1/2}(x)) = \frac{1}{2} \frac{\nabla_\alpha D_{\alpha}(x)}{D_\alpha^{1/2}(x)},$$
    %\item 
    $$\nabla_\theta((1-D_{\alpha,\theta}(z))^{1/2}) = - \frac{1}{2} \frac{\nabla_\theta D_{\alpha,\theta}(z)}{(1-D_{\alpha,\theta}(z))^{1/2}},$$
    %\item 
    $$\nabla_\alpha((1-D_{\alpha,\theta}(z))^{1/2}) = - \frac{1}{2} \frac{\nabla_\alpha D_{\alpha,\theta}(z)}{(1-D_{\alpha,\theta}(z))^{1/2}},$$
%\end{itemize}
the derivatives of first order of the loss function $HD^2_n(\theta,\alpha)$ are given by
\begin{align*}
    \nabla_\alpha \mathrm{HD}^2_n(\theta,\alpha) =& \int \nabla_\alpha D_\alpha(x) d\mu_n(x) - \int \nabla_\alpha D_{\alpha,\theta} (z)d\mu_g(z) \\
    & - \int \frac{\nabla_\alpha D_{\alpha}(x)}{D_{\alpha}(x)^{1/2}} d\mu_n(x) \int(1-D_{\alpha,\theta}(z))^{\frac{1}{2}} d\mu_g(z) \\
    & + \int D_\alpha^{\frac{1}{2}}(x) d\mu_n(x) \int\frac{\nabla_\alpha D_{\alpha,\theta}(z)}{(1-D_{\alpha,\theta}(z))^{1/2}} d\mu_g(z)
\end{align*}
and 
\begin{align*}
    \nabla_\theta \mathrm{HD}^2_n(\theta,\alpha) =&  - \int \nabla_\theta D_{\alpha,\theta} (z)d\mu_g(z) \\
    & + \int D_\alpha^{\frac{1}{2}}(x) d\mu_n(x) \int\frac{\nabla_\theta D_{\alpha,\theta}(z)}{(1-D_{\alpha,\theta}(z))^{1/2}} d\mu_g(z).
\end{align*}

\subsection{Derivatives of Second order}
\label{app:derivatives2}

Here, we report the the second derivatives of the Hellinger loss, which are used in the proof of Proposition \ref{prop2}.
The second derivatives with respect to $\theta$ are given as 
\begin{align*}
    \nabla_{\theta \theta^\top} \mathrm{HD}_n^2(\theta, \alpha) =& - \int \nabla_{\theta \theta^\top} D_{\alpha,\theta}(z)d\mu_g(z) \\
    & + \int D_{\alpha}^{\frac{1}{2}}(x) d\mu_n(x) \int \frac{\nabla_{\theta \theta^\top} D_{\alpha, \theta}(z)}{(1 - D_{\alpha, \theta}(z))^{\frac{1}{2}}} d\mu_g(z)\\
     & + \frac{1}{2} \int D_{\alpha}^{\frac{1}{2}}(x) d\mu_n(x) \int \frac{\nabla_\theta D_{\alpha, \theta}(z) \nabla_\theta^\top D_{\alpha, \theta}(z)}{(1 - D_{\alpha, \theta}(z))^{\frac{3}{2}}} d\mu_g(z)\\
     &= A_1 + A_2 + A_3, 
\end{align*}
and the second derivatives with respect to $\alpha$ are computed as
\begin{align*}
    \nabla_{\alpha \alpha^\top} \mathrm{HD}_n^2(\theta, \alpha) =& \int \nabla_{\alpha \alpha^\top} D_{\alpha}(x)d\mu_n(x) \\
    & - \int \nabla_{\alpha \alpha^\top} D_{\alpha, \theta}(z)d\mu_g(z) \\
    & - \int \frac{\nabla_{\alpha \alpha^\top} D_{\alpha}(x)}{D_{\alpha}^{\frac{1}{2}}(x)}d\mu_n(x) \int (1 - D_{\alpha, \theta}(z))^{\frac{1}{2}} d\mu_g(z)\\
    & +\frac{1}{2} \int \frac{\nabla_\alpha D_{\alpha}(x) \nabla_\alpha^\top D_{\alpha}(x)}{D_{\alpha}^{\frac{3}{2}}(x)}d\mu_n(x) \int (1 - D_{\alpha, \theta}(z))^{\frac{1}{2}} d\mu_g(z)\\
    & + \int \frac{\nabla_\alpha D_{\alpha}(x)}{D_{\alpha}^{\frac{1}{2}}(x)}d\mu_n(x) \int \frac{\nabla_\alpha D_{\alpha, \theta}(z)}{(1 - D_{\alpha, \theta}(z))^{\frac{1}{2}}} d\mu_g(z)\\
    & + \int D_{\alpha}^{\frac{1}{2}}(x) d\mu_n(x) \int \frac{\nabla_{\alpha \alpha^\top} D_{\alpha, \theta}(z)}{(1 - D_{\alpha, \theta}(z))^{\frac{1}{2}}} d\mu_g(z)\\
     & + \frac{1}{2} \int D_{\alpha}^{\frac{1}{2}}(x) d\mu_n(x) \int \frac{\nabla_\alpha D_{\alpha, \theta}(z) \nabla_\alpha^\top D_{\alpha, \theta}(z)}{(1 - D_{\alpha, \theta}(z))^{\frac{3}{2}}} d\mu_g(z)\\
     = & B_1 + B_2 + B_3 + B_4 + B_5+ B_6+ B_7,
\end{align*}
while the mixed derivatives are given by
\begin{align*}
    \nabla_{\theta \alpha} \mathrm{HD}_n^2(\theta, \alpha) =& - \int \nabla_{\theta \alpha}  D_{\alpha,\theta}(z)d\mu_g(z) \\
    & + \frac{1}{2} \int \frac{\nabla_\alpha D_{\alpha}(x)}{D_{\alpha}^{\frac{1}{2}}(x)} d\mu_n(x) \int \frac{\nabla_\theta D_{\alpha,\theta}(z)}{(1 - D_{\alpha, \theta}(z))^{\frac{1}{2}}} d\mu_g(z) \\
    & + \int D_{\alpha}^{\frac{1}{2}}(x) d\mu_n(x) \int \frac{\nabla_{\theta \alpha} D_{\alpha, \theta}(z)}{(1 - D_{\alpha, \theta}(z))^{\frac{1}{2}}} d\mu_g(z)\\
     & + \frac{1}{2} \int D_{\alpha}^{\frac{1}{2}}(x) d\mu_n(x) \int \frac{\nabla_\alpha D_{\alpha, \theta}(z) \nabla_\theta^\top D_{\alpha, \theta}(z)}{(1 - D_{\alpha, \theta}(z))^{\frac{3}{2}}} d\mu_g(z)\\
     = & C_1 + C_2 + C_3 + C_4.
\end{align*}
Notice that, the term $\nabla_{\alpha \theta} \mathrm{HD}_n^2$ differs from  $\nabla_{\theta \alpha} \mathrm{HD}_n^2$ computed above, only for the order of derivation in $C_1$ and $C_3$.
\end{appendix}

\bibliography{gan}



\section*{Supplementary material}

# Supplementary Material: "Hellinger loss function for Generative Adversarial Networks"

Giovanni Saraceno*[1], Anand N. Vidyashankar[2], and Claudio Agostinelli[3]

[1]Department of Statistical Sciences, University of Padova, Italy
[2]Department of Statistics, George Mason University, VA, USA
[3]Department of Mathematics, University of Trento, Italy

## S1  Kernel density estimator $\mu_n^{\text{KDE}}$

As pointed out in Remark 3.1 of the manuscript, the empirical measure $\mu_n$ is not the unique choice of sample-based estimator of $\mu_*$. One possible alternative is to approximate $\mu_*$ by a smoothed estimator such as the kernel density estimator (KDE)

$$d\mu_n^{\text{KDE}}(x) = h_n(x)d\mu(x), \qquad h_n(x) = \frac{1}{nc_n^d}\sum_{i=1}^n K\left(\frac{x - X_i}{c_n}\right)$$

with kernel function $K$ and bandwidth parameter $c_n$. Considering the following additional assumptions, we can state analogous theorems for the consistency and asymptotic normality of $(\theta_n, \alpha_n)$.

($H_K$) The kernel function $K$ is such that $K(x) \to 0$ as $|x| \to \infty$, $\int |K(x)|dx < \infty$.

($H_{n,1}$) $c_n \to 0$ and $nc_n^d \to \infty$ as $n \to \infty$.

($H_{n,2}$) The parameter $c_n$ satisfies $\sqrt{n}c_n^2 \to 0$ as $n \to \infty$.

First, we introduce the following Lemma which is a modified version of Lemma 3.1 of the manuscript when the KDE $\mu_n^{\text{KDE}}$ is considered.

**Lemma S1.1.** *Consider $X_1, \ldots, X_n$ i.i.d. observations such that $X_i \sim p_*$ and let $h_n(x)$ be the kernel density estimator of $p_*$. Let $f$ be a d-dimensional function uniformly bounded. Suppose that assumptions ($H_K$) and ($H_{n,1}$) hold, then, as $n \to \infty$,*

*(i)*
$$\sup_x |h_n(x) - p_*(x)| \to 0$$

*in probability and a.s.*

---

*giovanni.saraceno@unipd.it

*(ii)*
$$\int f(x)(h_n(x) - p_*(x))d\mu(x) \longrightarrow 0$$
*in probability;*

*(iii)*
$$\sqrt{n}\int f(x)((h_n(x) - p_*(x)))d\mu(x) \longrightarrow N_d(0, \Sigma_f)$$
*in distribution, where* $\Sigma_f = Var[f(X)]\int[K(x)]^2 dx$.

*Proof.* **Part (i)** The result follows from Glick's theorem [Glick, 1974].
**Part (ii)** Given the assumptions $(H_{n,1})$, $h_n(x)$ is a consistent estimator of $p_*(x)$. Since $f(x)$ is uniformly bounded, by the dominated convergence theorem
$$\int |f(x)(h_n(x) - p_*(x))|\, d\mu(x)$$
$$\leq \sup_x |f(x)| \int |h_n(x) - p_*(x)|\, d\mu(x) \to 0 \quad \text{as } n \to \infty.$$
This implies that the integral itself tends to 0 in probability.
**Part (iii)** Consider the term
$$T_n = \sqrt{n}\int f(x)(h_n(x) - p_*(x))d\mu(x).$$
Substituting $h_n(x)$ into the expression for $T_n$, we get
$$T_n = \frac{1}{\sqrt{n}}\sum_{i=1}^n \int f(x)\left(\frac{1}{c_n^d}K\left(\frac{x - X_i}{c_n}\right) - p_*(x)\right) d\mu(x) = \frac{1}{\sqrt{n}}\sum_{i=1}^n Y_i.$$
Note that $h_n(x)$ is asymptotically unbiased, then, as $c_n \to 0$, $\mathbb{E}[Y_i] \to 0$ and
$$\text{Var}(Y_i) \to \Sigma_f = \text{Var}(f(X))\int[K(x)]^2 dx.$$
Then, by the central limit theorem
$$T_n \xrightarrow{d} N(0, \Sigma_f).$$
□

With this Lemma, we can prove the consistency property of the Hellinger Loss with KDE.

**Theorem S1.1** (Consistency). *If* $(H_D)$, $(H_G)$, $(H_K)$ *and* $(H_{n,1})$ *hold, then*
$$(\theta_n, \alpha_n) \xrightarrow{a.s.} (\theta_*, \alpha_*) \text{ as } n \to \infty.$$

*Proof.* The proof follows the same steps as in the proof of Theorem 3.1 of the manuscript. In the first step we show that $HD_n^2(\theta, \alpha)$ converges uniformly, almost surely, to $HD^2(\theta, \alpha)$. Note that $D_\alpha(x) \leq 1$, then
$$|h_{1,n}(\alpha) - h_1(\alpha)| = \left|\int D_\alpha(x)h_n(x)d\mu(x) - \int D_\alpha(x)p_*(x)d\mu(x)\right|$$
$$= \left|\int D_\alpha(x)(h_n(x) - p_*(x))d\mu(x)\right|$$
$$\leq \int |h_n(x) - p_*(x)|d\mu(x),$$

then
$$\lim_{n\to\infty} \int |h_n(x) - p_*(x)| d\mu(x) = 0 \quad \text{a.s.}$$

by point $(i)$ of Lemma S1.1. In a similar way, we have
$$0 \le \limsup_{n\to\infty} \sup_{(\theta,\alpha)\in\Theta\times\Lambda} |\gamma_n(\theta,\alpha) - \gamma(\theta,\alpha)| = 0.$$

The second step is the same as that of Theorem 3.1. $\square$

Finally, we introduce the following propositions to prove the analogous result of asymptotic normality.

**Proposition S1.1.** *Assume that assumptions $(H_K)$, $(H_{n,1})$, $(H_{n,2})$ and $(H_1) - (H_3)$ are satisfied. Then*
$$\sqrt{n}\nabla \mathrm{HD}_n^2(\theta_*, \alpha_*) \longrightarrow N(0, S^{\mathrm{KDE}}),$$
*where $S^{\mathrm{KDE}}$ is a non-singular covariance matrix.*

*Proof.* Recall that $D_\alpha(x)$ is bounded for each $\alpha$. Then by point $(ii)$ of Lemma S1.1
$$\int D_\alpha(x)(h_n(x) - p_*(x))d\mu(x) \longrightarrow 0,$$
$$\int D_\alpha^{\frac{1}{2}}(x)(h_n(x) - p_*(x))d\mu(x) \longrightarrow 0 \text{ as } n \to \infty.$$

The derivative with respect to $\theta$ given in Appendix A.1 computed at $(\theta_*, \alpha_*)$ can be rewritten as
$$\sqrt{n}\nabla_\theta HD_n^2(\theta_*, \alpha_*) =$$
$$- \sqrt{n} \int \nabla_\theta D_{\alpha_*,\theta_*}(z) g(z) d\mu(z)$$
$$+ \sqrt{n} \int D_{\alpha_*}^{\frac{1}{2}}(x)(h_n(x) - p_*(x))d\mu(x) \int \frac{\nabla_\theta D_{\alpha_*,\theta_*}(z)}{(1 - D_{\alpha_*,\theta_*}(z))^{1/2}} g(z) d\mu(z)$$
$$+ \sqrt{n} \int D_{\alpha_*}^{\frac{1}{2}}(x) p_*(x) d\mu(x) \int \frac{\nabla_\theta D_{\alpha_*,\theta_*}(z)}{(1 - D_{\alpha_*,\theta_*}(z))^{1/2}} g(z) d\mu(z)$$
$$= T_1 + T_{2n} + T_3.$$

By Lemma S1.1, we have that
$$T_{2n} \xrightarrow{d} N(0, S_1^{\mathrm{KDE}}),$$

with
$$S_1^{\mathrm{KDE}} = Var(\Delta_1(X)) \int [K(x)]^2 d(x) \quad \text{and} \quad \Delta_1(X) = D_{\alpha_*}^{\frac{1}{2}}(X) \int \frac{\nabla_\theta D_{\alpha_*,\theta_*}(z)}{(1 - D_{\alpha_*,\theta_*}(z))^{1/2}} g(z) d\mu(z).$$

Notice that $T_1 + T_3 = 0$ since it corresponds to $\nabla_\theta HD^2(\theta_*, \alpha_*) = 0$.

3

Let us now consider the derivative with respect to $\alpha$ given in Appendix A.1 computed at $(\theta_*, \alpha_*)$, that can be rewritten as

$$\sqrt{n}\nabla_\alpha HD_n^2(\theta_*, \alpha_*) = \sqrt{n}\int \nabla_\alpha D_{\alpha_*}(x)(h_n(x) - p_*(x))d\mu(x)$$
$$+ \sqrt{n}\int \nabla_\alpha D_{\alpha_*}(x)p_*(x)d\mu(x)$$
$$- \sqrt{n}\int \nabla_\alpha D_{\alpha_*,\theta_*}(z)g(z)d\mu(z)$$
$$- \sqrt{n}\int \frac{\nabla_\alpha D_{\alpha_*}(x)}{D_{\alpha_*}(x)^{1/2}}(h_n(x) - p_*(x))d\mu(x)\int (1 - D_{\alpha_*,\theta_*}(z))^{\frac{1}{2}}g(z)d\mu(z)$$
$$- \sqrt{n}\int \frac{\nabla_\alpha D_{\alpha_*}(x)}{D_{\alpha_*}(x)^{1/2}}p_*(x)d\mu(x)\int (1 - D_{\alpha_*,\theta_*}(z))^{\frac{1}{2}}g(z)d\mu(z)$$
$$+ \sqrt{n}\int D_{\alpha_*}^{\frac{1}{2}}(x)(h_n(x) - p_*(x))d\mu(x)\int \frac{\nabla_\alpha D_{\alpha_*,\theta_*}(z)}{(1 - D_{\alpha_*,\theta_*}(z))^{1/2}}g(z)d\mu(z)$$
$$+ \sqrt{n}\int D_{\alpha_*}^{\frac{1}{2}}(x)p_*(x)d\mu(x)\int \frac{\nabla_\alpha D_{\alpha_*,\theta_*}(z)}{(1 - D_{\alpha_*,\theta_*}(z))^{1/2}}g(z)d\mu(z)$$
$$= J_{1,n} + J_2 + J_3 + J_{4,n} + J_5 + J_{6,n} + J_7.$$

Notice that $J_2 + J_3 + J_5 + J_7 = 0$ since it corresponds to $\nabla_\alpha HD^2(\theta_*, \alpha_*) = 0$. The remaining terms can be rewritten as

$$J_{1,n} + J_{4,n} + J_{6,n} = \sqrt{n}\int \Delta_2(x)(h_n(x) - p_*(x))d\mu(x)$$

where

$$\Delta_2(x) = \nabla_\alpha D_{\alpha_*}(x) - \frac{\nabla_\alpha D_{\alpha_*}(x)}{D_{\alpha_*}(x)^{1/2}}\int (1 - D_{\alpha_*,\theta_*}(z))^{\frac{1}{2}}g(z)d\mu(z)$$
$$+ D_{\alpha_*}^{\frac{1}{2}}(x)\int \frac{\nabla_\alpha D_{\alpha_*,\theta_*}(z)}{(1 - D_{\alpha_*,\theta_*}(z))^{1/2}}g(z)d\mu(z)$$

By Lemma S1.1, we have that

$$J_{1,n} + J_{4,n} + J_{6,n} \xrightarrow{d} N(0, S_2^{\text{KDE}}) \text{ with } S_2^{\text{KDE}} = Var(\Delta_2(X))\int [K(x)]^2 dx.$$

Then by the central limit theorem and the continuous mapping theorem, we have that

$$(T_{2n}; J_{1n} + J_{4n} + J_{6n}) \longrightarrow N(0, S^{\text{KDE}})$$

where $S^{\text{KDE}} = \begin{pmatrix} S_1^{\text{KDE}} & S_{12}^{\text{KDE}} \\ S_{21}^{\text{KDE}} & S_2^{\text{KDE}} \end{pmatrix}$ where $S_{12}^{\text{KDE}} = \text{Cov}(\Delta_1(X), \Delta_2(X))$. □

## S2 Proof of Proposition 3.2

In this section, we provide additional details about the convergence of the other terms of the second derivatives of the Hellinger loss considered in Proposition 3.2. In the proof of Proposition 3.2 we showed the convergence of the second derivatives with respect to $\theta$.

Let's now consider the second derivatives with respect to $\alpha$ reported in Appendix A.2. Using simple operations, it can be rewritten as

$$B_3 + B_4 + B_5 + B_6 + B_7 = \int \Delta_B(x) \left(d\mu_n(x) - d\mu_*(x)\right)$$
$$+ \int \Delta_B(x) d\mu_*(x)$$

with

$$\Delta_B(x) = -\frac{\nabla_{\alpha\alpha^\top} D_\alpha(x)}{D_\alpha^{\frac{1}{2}}(x)} \int (1 - D_{\alpha,\theta}(z))^{\frac{1}{2}} d\mu_g(z)$$
$$+ \frac{1}{2} \frac{\nabla_\alpha D_\alpha(x) \nabla_\alpha^\top D_\alpha(x)}{D_\alpha^{\frac{3}{2}}(x)} \int (1 - D_{\alpha,\theta}(z))^{\frac{1}{2}} d\mu_g(z)$$
$$+ \frac{\nabla_\alpha D_\alpha(x)}{D_\alpha^{\frac{1}{2}}(x)} \int \frac{\nabla_\alpha D_{\alpha,\theta}(z)}{(1 - D_{\alpha,\theta}(z))^{\frac{1}{2}}} d\mu_g(z)$$
$$+ D_\alpha^{\frac{1}{2}}(x) \int \frac{\nabla_{\alpha\alpha^\top} D_{\alpha,\theta}(z)}{(1 - D_{\alpha,\theta}(z))^{\frac{1}{2}}} d\mu_g(z)$$
$$+ \frac{1}{2} D_\alpha^{\frac{1}{2}}(x) \int \frac{\nabla_\alpha D_{\alpha,\theta}(z) \nabla_\alpha^\top D_{\alpha,\theta}(z)}{(1 - D_{\alpha,\theta}(z))^{\frac{3}{2}}} d\mu_g(z).$$

By assumption $(H_2)$, $\Delta_B(x)$ is bounded and continuous, then the first term converges to 0 as $n \to \infty$ by point $(i)$ of Lemma 3.1. Finally, we consider the term of mixed derivatives, which can be rewritten as

$$C_2 + C_3 + C_4 = \int \Delta_C(x) \left(d\mu_n(x) - d\mu_*(x)\right)$$
$$+ \int \Delta_C(x) d\mu_*(x)$$

with

$$\Delta_C(x) = \frac{1}{2} \frac{\nabla_\alpha D_\alpha(x)}{D_\alpha^{\frac{1}{2}}(x)} \int \frac{\nabla_\theta D_{\alpha,\theta}(z)}{(1 - D_{\alpha,\theta}(z))^{\frac{1}{2}}} d\mu_g(z)$$
$$+ D_\alpha^{\frac{1}{2}}(x) \int \frac{\nabla_{\theta\alpha} D_{\alpha,\theta}(z)}{(1 - D_{\alpha,\theta}(z))^{\frac{1}{2}}} d\mu_g(z)$$
$$+ \frac{1}{2} D_\alpha^{\frac{1}{2}}(x) \int \frac{\nabla_\alpha D_{\alpha,\theta}(z) \nabla_\theta^\top D_{\alpha,\theta}(z)}{(1 - D_{\alpha,\theta}(z))^{\frac{3}{2}}} d\mu_g(z).$$

By Lemma 3.1 the first term converges to zero.

## S3   Influence Function for the standard loss function

In this section, we provide the computation of the influence function for the standard loss function used in GANs.

Let $p_\varepsilon(x) = (1 - \varepsilon) q_{\theta_0}(x) + \varepsilon h(x)$ denote the contaminated density at $x$, with the contaminating distribution $h(x)$ and $\theta_*$ denote the parameter values for which the model density $q_{\theta_*}$ coincides with the true distribution $p_*$. The underlying contaminated random

vector $Z \sim g$ is such that $X = G_{\theta_*}(Z)$, and for every $\varepsilon$ we have $X \sim p_\varepsilon(x)$ the contaminated random vector that generates the data. We study the standard GAN log-loss function under contamination given as

$$L_\varepsilon(\theta, \alpha) = \int \ln(D_\alpha(x)) p_\varepsilon(x) d\mu(x) + \int \ln(1 - D_\alpha(x)) q_\theta(x) d\mu(x).$$

For all $\varepsilon \in [0, 1)$, we define

$$(\theta_\varepsilon, \alpha_\varepsilon) = \arg \inf_\alpha \sup_\theta L_\varepsilon(\theta, \alpha).$$

Notice that for $\varepsilon = 0$, $L_0(\theta, \alpha)$ denotes the uncontaminated objective function. Then, the influence functions for the standard loss is defined as

$$(IF(\theta), IF(\alpha)) = \frac{\partial}{\partial \varepsilon}(\theta_\varepsilon, \alpha_\varepsilon)\bigg|_{\varepsilon=0}.$$

As similarly done in Section 4 of the manuscript, the optimizer $(\theta_\varepsilon, \alpha_\varepsilon)$ is solution of $\nabla L_\varepsilon(\theta_\varepsilon, \alpha_\varepsilon) = 0$, for a fixed $\varepsilon \in [0, 1)$, hence, in order to compute the influence function we can consider, for all $\varepsilon \in [0, 1)$

$$\frac{\partial}{\partial \varepsilon} \nabla L_\varepsilon(\theta_\varepsilon, \alpha_\varepsilon) = 0 .$$

We have

$$\nabla_\alpha L_\varepsilon(\alpha, \theta) = \int \frac{\nabla_\alpha D_\alpha(x)}{D_\alpha(x)} p_\varepsilon(x) d\mu(x) - \int \frac{\nabla_\alpha D_\alpha(x)}{1 - D_\alpha(x)} q_\theta(x) d\mu(x)$$
$$= A_\alpha + B_\alpha$$

and

$$\nabla_\theta L_\varepsilon(\alpha, \theta) = \int \ln(1 - D_\alpha(x)) s_\theta(x) q_\theta(x) d\mu(x) = C_\theta ,$$

where $s_\theta(x) = \nabla_\theta \log q_\theta(x)$ is the usual score function. We need to compute the derivatives with respect to $\varepsilon$ of these terms and evaluate the expressions at $\varepsilon = 0$. For the first term

$$\frac{\partial}{\partial \varepsilon} A_\alpha = \int \frac{\frac{\partial}{\partial \varepsilon} \nabla_\alpha D_{\alpha_\varepsilon}(x)}{D_{\alpha_\varepsilon}(x)} p_\varepsilon(x) d\mu(x) - \int \frac{\nabla_\alpha D_{\alpha_\varepsilon}(x) \frac{\partial}{\partial \varepsilon} D_{\alpha_\varepsilon}(x)}{D_{\alpha_\varepsilon}^2(x)} p_\varepsilon(x) d\mu(x)$$
$$+ \int \frac{\nabla_\alpha D_{\alpha_\varepsilon}(x)}{D_{\alpha_\varepsilon}(x)} \frac{\partial}{\partial \varepsilon} p_\varepsilon(x) d\mu(x)$$
$$= \int \frac{\nabla_{\alpha\alpha^\top} D_{\alpha_\varepsilon}(x)}{D_{\alpha_\varepsilon}(x)} p_\varepsilon(x) d\mu(x) \frac{\partial}{\partial \varepsilon} \alpha_\varepsilon + \int \frac{\nabla_\alpha D_{\alpha_\varepsilon}(x) \nabla_\alpha^\top D_{\alpha_\varepsilon}(x)}{D_{\alpha_\varepsilon}^2(x)} p_\varepsilon(x) d\mu(x) \frac{\partial}{\partial \varepsilon} \alpha_\varepsilon$$
$$+ \int \frac{\nabla_\alpha D_{\alpha_\varepsilon}(x)}{D_{\alpha_\varepsilon}(x)} (h(x) - q_{\theta_0}(x)) d\mu(x)$$

with $\varepsilon = 0$ we obtain
$$= A_1 \, \text{IF}(\alpha) + A_2 \, \text{IF}(\alpha) + A_3$$

6

Following similar steps, the derivatives computed at $\varepsilon = 0$ are given as

$$\frac{\partial}{\partial \varepsilon} B_\alpha = -\int \frac{\frac{\partial}{\partial \varepsilon}\nabla_\alpha D_{\alpha_\varepsilon}(x)}{1 - D_{\alpha_\varepsilon}(x)} q_{\theta_\varepsilon}(x) d\mu(x) + \int \frac{\nabla_\alpha D_{\alpha_\varepsilon}(x) \frac{\partial}{\partial \varepsilon}(1 - D_{\alpha_\varepsilon}(x))}{(1 - D_{\alpha_\varepsilon}(x))^2} q_{\theta_\varepsilon}(x) d\mu(x)$$

$$- \int \frac{\nabla_\alpha D_{\alpha_\varepsilon}(x)}{1 - D_{\alpha_\varepsilon}(x)} \frac{\partial}{\partial \varepsilon} q_{\theta_\varepsilon}(x) d\mu(x)$$

$$= -\int \frac{\nabla_{\alpha\alpha^\top} D_{\alpha_\varepsilon}(x)}{1 - D_{\alpha_\varepsilon}(x)} q_{\theta_\varepsilon}(x) d\mu(x) \frac{\partial}{\partial \varepsilon}\alpha_\varepsilon - \int \frac{\nabla_\alpha D_{\alpha_\varepsilon}(x) \nabla_\alpha^\top D_{\alpha_\varepsilon}(x)}{(1 - D_{\alpha_\varepsilon}(x))^2} q_{\theta_\varepsilon}(x) d\mu(x) \frac{\partial}{\partial \varepsilon}\alpha_\varepsilon$$

$$- \int \frac{\nabla_\alpha D_{\alpha_\varepsilon}(x)}{1 - D_{\alpha_\varepsilon}(x)} s_{\theta_\varepsilon}(x) q_{\theta_\varepsilon}(x) d\mu(x) \frac{\partial}{\partial \varepsilon}\theta_\varepsilon$$

with $\varepsilon = 0$ we obtain

$$= B_1 \operatorname{IF}(\alpha) + B_2 \operatorname{IF}(\alpha) + B_3 \operatorname{IF}(\theta)$$

$$\frac{\partial}{\partial \varepsilon} C_\theta = -\int \frac{\frac{\partial}{\partial \varepsilon} D_{\alpha_\varepsilon}(x)}{1 - D_{\alpha_\varepsilon}(x)} s_{\theta_\varepsilon}(x) q_{\theta_\varepsilon}(x) d\mu(x) + \int \ln(1 - D_{\alpha_\varepsilon}(x)) \frac{\partial}{\partial \varepsilon} s_{\theta_\varepsilon}(x) q_{\theta_\varepsilon}(x) d\mu(x)$$

$$+ \int \ln(1 - D_{\alpha_\varepsilon}(x)) s_{\theta_\varepsilon}(x) \frac{\partial}{\partial \varepsilon} q_{\theta_\varepsilon}(x) d\mu(x)$$

$$= -\int \frac{\nabla_{\alpha_\varepsilon} D_{\alpha_\varepsilon}(x)}{1 - D_{\alpha_\varepsilon}(x)} s_{\theta_\varepsilon}(x) q_{\theta_\varepsilon}(x) d\mu(x) \frac{\partial}{\partial \varepsilon}\alpha_\varepsilon$$

$$+ \int \ln(1 - D_{\alpha_\varepsilon}(x)) \frac{\nabla_{\theta,\theta^\top} q_\theta(x)}{q_{\theta_\varepsilon}(x)} q_{\theta_\varepsilon}(x) d\mu(x) \frac{\partial}{\partial \varepsilon}\theta_\varepsilon$$

$$= C_1 \operatorname{IF}(\alpha) + C_2 \operatorname{IF}(\theta) \ .$$

Combining all the terms together, we have the following two equations

$$0 = A_3 + [A_1 + A_2 + B_1 + B_2] IF(\alpha) + B_3 \operatorname{IF}(\theta)$$
$$= D_0 + D_\alpha \operatorname{IF}(\alpha) + D_\theta \operatorname{IF}(\theta),$$

and

$$0 = C_1 IF(\alpha) + C_2 \operatorname{IF}(\theta)$$

Solving the system we have

$$\operatorname{IF}(\alpha) = -(D_\alpha - D_\theta C_\theta^{-1} C_\alpha)^{-1} D_0$$
$$\operatorname{IF}(\theta) = -C_\theta^{-1} C_\alpha \operatorname{IF}(\alpha) = C_\theta^{-1} C_\alpha (D_\alpha - D_\theta C_\theta^{-1} C_\alpha)^{-1} D_0 \ .$$

## S4  Profiled Hellinger Distance Estimator

In this section, we develop the asymptotic theory for the profiled Hellinger distance estimator obtained by optimizing the generator parameter after profiling out the discriminator. We first establish the continuity of the empirical and population profiled criteria and the existence of optimizers, and show that profiling preserves the uniform convergence of $\operatorname{HD}_n^2(\theta, \alpha)$ with its population counterpart. We derive a central limit theorem for the profiled estimator and relate it to the joint CLT for the full estimator. These results also extend to the approximated objective.

**Lemma S4.1** (Supremum inequality). *Let $(A, \mathcal{A})$ be an index set and let $u, v : A \to \mathbb{R}$. Then*
$$\left|\sup_{a \in A} u(a) - \sup_{a \in A} v(a)\right| \leq \sup_{a \in A} \left|u(a) - v(a)\right|.$$

*Proof.* Fix $\varepsilon > 0$ and choose $a_\varepsilon \in A$ such that $\sup_a u(a) \leq u(a_\varepsilon) + \varepsilon$. Then
$$\sup_a u(a) - \sup_a v(a) \leq u(a_\varepsilon) - \sup_a v(a) + \varepsilon \leq u(a_\varepsilon) - v(a_\varepsilon) + \varepsilon \leq \sup_a |u(a) - v(a)| + \varepsilon.$$

Letting $\varepsilon \downarrow 0$ yields $\sup_a u(a) - \sup_a v(a) \leq \sup_a |u(a) - v(a)|$. Interchanging $u$ and $v$ gives the reverse inequality and hence the claim. □

**Lemma S4.2** (Continuity and existence of optimizers for the profiled criteria). *Assume $(H_D)$ and $(H_G)$. For each $n \in \mathbb{N} \cup \{\infty\}$, define*
$$S_n(\theta) := \sup_{\alpha \in \Lambda} \mathrm{HD}_n^2(\theta, \alpha), \qquad \theta \in \Theta,$$

*Then:*

(i) $\mathrm{HD}_n^2$ *is continuous on $\Theta \times \Lambda$ for each $n$.*

(ii) $S_n$ *is continuous on $\Theta$ for each $n$.*

(iii) *For each $\theta \in \Theta$, the supremum in $S_n(\theta)$ is attained: there exists $\alpha_n(\theta) \in \Lambda$ such that $S_n(\theta) = \mathrm{HD}_n^2(\theta, \alpha_n(\theta))$.*

*Proof.* (i) Under $(H_D)$ and $(H_G)$, the maps $(x, \alpha) \mapsto D_\alpha(x)$ and $(z, \theta, \alpha) \mapsto D_\alpha(G_\theta(z))$ are continuous and bounded by 1. Inspecting the decomposition
$$\mathrm{HD}_n^2(\theta, \alpha) = h_{1,n}(\alpha) + h_2(\theta, \alpha) - 2\gamma_n(\theta, \alpha),$$

with
$$h_{1,n}(\alpha) = \int D_\alpha(x)\, h_n(x)\, d\mu(x),$$
$$h_2(\theta, \alpha) = \int \bigl(1 - D_\alpha(G_\theta(z))\bigr)\, g(z)\, d\mu(z),$$
$$\gamma_n(\theta, \alpha) = \left(\int D_\alpha^{1/2}(x)\, h_n(x)\, d\mu(x)\right)\left(\int \bigl(1 - D_\alpha(G_\theta(z))\bigr)^{1/2} g(z)\, d\mu(z)\right),$$

we see that continuity of $\mathrm{HD}_n^2$ follows in each term by dominated convergence, using the uniform bound $0 \leq D_\alpha \leq 1$ and the continuity of the integrands in $(\theta, \alpha)$. (The case $n = \infty$ is identical with $h_n$ replaced by $p^*$.)

(ii) Since $\mathrm{HD}_n^2$ is continuous on the compact set $\Theta \times \Lambda$, it is uniformly continuous. Thus, for every $\varepsilon > 0$ there exists $\delta > 0$ such that $|\mathrm{HD}_n^2(\theta, \alpha) - \mathrm{HD}_n^2(\theta', \alpha)| < \varepsilon$ whenever $\|\theta - \theta'\| < \delta$, for all $\alpha \in \Lambda$. Taking suprema over $\alpha$ yields $|S_n(\theta) - S_n(\theta')| \leq \varepsilon$, proving continuity of $S_n$.

(iii) By (ii) and compactness of $\Lambda$, Weierstrass' theorem gives $\alpha_n(\theta) \in \arg\max_{\alpha \in \Lambda} \mathrm{HD}_n^2(\theta, \alpha)$. □

**Assumption S4.1** ($H_n$). *The conditions of Theorem 3.2 in the main paper hold, so that*
$$\sup_{(\theta, \alpha) \in \Theta \times \Lambda} \left|\mathrm{HD}_n^2(\theta, \alpha) - HD^2(\theta, \alpha)\right| \longrightarrow 0 \quad \textit{almost surely as } n \to \infty.$$

**Proposition S4.1** (Uniform profiling over the discriminator). *Assume $(H_n)$. Then*

$$\sup_{\theta \in \Theta} \left| S_n(\theta) - S(\theta) \right| \xrightarrow{a.s.} 0,$$

*where*

$$S_n(\theta) = \sup_{\alpha \in \Lambda} \mathrm{HD}_n^2(\theta, \alpha), \qquad S(\theta) = \sup_{\alpha \in \Lambda} \mathrm{HD}^2(\theta, \alpha).$$

*Proof.* Fix $\theta \in \Theta$ and apply Lemma S4.1 with $u(\alpha) = \mathrm{HD}_n^2(\theta, \alpha)$ and $v(\alpha) = \mathrm{HD}^2(\theta, \alpha)$ to get

$$\left| S_n(\theta) - S(\theta) \right| \leq \sup_{\alpha \in \Lambda} \left| \mathrm{HD}_n^2(\theta, \alpha) - \mathrm{HD}^2(\theta, \alpha) \right|.$$

Taking the supremum over $\theta \in \Theta$ yields

$$\sup_{\theta \in \Theta} \left| S_n(\theta) - S(\theta) \right| \leq \sup_{(\theta, \alpha) \in \Theta \times \Lambda} \left| \mathrm{HD}_n^2(\theta, \alpha) - \mathrm{HD}^2(\theta, \alpha) \right|.$$

By the first part of Theorem 3.2, the right-hand side converges to 0 almost surely under $(H_n)$. Hence the claim follows. □

**Theorem S4.1** (Consistency of the profiled estimator). *Assume $(H_D)$, $(H_G)$, and $(H_n)$. Let*

$$\hat{\theta}_n \in \arg\min_{\theta \in \Theta} S_n(\theta), \qquad \theta^* \in \arg\min_{\theta \in \Theta} S(\theta).$$

*If $\Theta$ is compact and $\theta^*$ is the unique minimizer of $S$, then $\hat{\theta}_n \to \theta^*$ almost surely.*

*Proof.* By Lemma S4.2(ii) and compactness of $\Theta$, the minima of $S_n$ and $S$ are attained. Proposition S4.1 gives $\sup_{\theta \in \Theta} |S_n(\theta) - S(\theta)| \to 0$ a.s.

*Step 1 (modulus of separation of the minimum).* We claim that for each $r > 0$ there exists $\eta(r) > 0$ such that

$$\inf\{S(\theta) : \theta \in \Theta, \ \|\theta - \theta^*\| \geq r\} \geq S(\theta^*) + \eta(r).$$

If not, we can find a sequence $(\theta_k) \subset \Theta$ with $\|\theta_k - \theta^*\| \geq r$ and $S(\theta_k) \downarrow S(\theta^*)$. By compactness of $\Theta$ there is a convergent subsequence $\theta_{k_\ell} \to \bar{\theta}$ with $\|\bar{\theta} - \theta^*\| \geq r$. Continuity of $S$ (Lemma S4.2(ii) with $n = \infty$) gives $S(\bar{\theta}) = \lim_\ell S(\theta_{k_\ell}) = S(\theta^*)$, contradicting uniqueness of $\theta^*$.

*Step 2 (contradiction argument).* Fix $r > 0$ and set $\eta = \eta(r) > 0$ from Step 1. Almost surely, for $n$ large enough we have $\sup_\theta |S_n(\theta) - S(\theta)| < \eta/2$. Then

$$S_n(\hat{\theta}_n) \leq S_n(\theta^*) \leq S(\theta^*) + \eta/2. \tag{1}$$

If $\|\hat{\theta}_n - \theta^*\| \geq r$, then Step 1 implies $S(\hat{\theta}_n) \geq S(\theta^*) + \eta$, hence

$$S_n(\hat{\theta}_n) \geq S(\hat{\theta}_n) - \eta/2 \geq S(\theta^*) + \eta/2,$$

which contradicts the previous bound. Therefore $\|\hat{\theta}_n - \theta^*\| < r$ eventually almost surely. Since $r > 0$ is arbitrary, we conclude $\hat{\theta}_n \to \theta^*$ almost surely. □

**Remark S4.1** (Empirical and population optimizers are linked)**.** For each $n$, Lemma S4.2(iii) gives
$$\alpha_n(\theta) \in \arg\max_{\alpha \in \Lambda} \mathrm{HD}_n^2(\theta, \alpha),$$
so any saddle-point solution $(\theta_n, \alpha_n) \in \arg\min_{\theta \in \Theta} \arg\max_{\alpha \in \Lambda} \mathrm{HD}_n^2(\theta, \alpha)$ satisfies
$$\theta_n \in \arg\min_{\theta \in \Theta} S_n(\theta).$$

Theorem S4.1 therefore implies the almost sure consistency of the generator estimate obtained by profiling over the discriminator parameter.

**Proposition S4.2** (approximated objective)**.** *Define*
$$\widetilde{S}_n(\theta) := \sup_{\alpha \in \Lambda} \widetilde{L}_n(\theta, \alpha)$$
*and*
$$\widetilde{S}(\theta) := \sup_{\alpha \in \Lambda} \widetilde{L}(\theta, \alpha),$$
*where $\widetilde{L}_n(\theta, \alpha) = 2\gamma_n(\theta, \alpha) - 2$ and $\widetilde{L}(\theta, \alpha) = 2\gamma(\theta, \alpha) - 2$. Under $(H_n)$, $\sup_\theta |\widetilde{S}_n(\theta) - \widetilde{S}(\theta)| \to 0$ almost surely. If, in addition, $\Theta$ is compact and $\widetilde{S}$ has a unique minimizer $\theta^\dagger$, then any $\hat{\theta}_n^{\mathrm{approx}} \in \arg\min_\Theta \widetilde{S}_n(\theta)$ satisfies $\hat{\theta}_n^{\mathrm{approx}} \to \theta^\dagger$ almost surely.*

*Proof.* For fixed $\theta$, Lemma S4.1 gives $\left|\sup_\alpha \gamma_n(\theta, \alpha) - \sup_\alpha \gamma(\theta, \alpha)\right| \leq \sup_\alpha |\gamma_n(\theta, \alpha) - \gamma(\theta, \alpha)|$. Taking $\sup_\theta$ and using the a.s. uniform convergence $\sup_{\theta,\alpha} |\gamma_n(\theta, \alpha) - \gamma(\theta, \alpha)| \to 0$ (from the proof of Theorem 3.2) yields the first claim. The consistency claim follows exactly as in Theorem S4.1. □

## S4.1 Central Limit Theorem

**Definition S4.1** (Profiled objective and profiled estimator)**.** For each $\theta \in \Theta$, let
$$S_n(\theta) := \sup_{\alpha \in \Lambda} \mathrm{HD}_n^2(\theta, \alpha), \quad S(\theta) := \sup_{\alpha \in \Lambda} \mathrm{HD}^2(\theta, \alpha).$$

A *profiled* generator estimator is any
$$\hat{\theta}_n \in \arg\min_{\theta \in \Theta} S_n(\theta),$$
with a corresponding $\hat{\alpha}_n \in \arg\max_{\alpha \in \Lambda} \mathrm{HD}_n^2(\hat{\theta}_n, \alpha)$.

The next assumption concerns block nonsigularity and interior solutions.

**Assumption S4.2.** *(i) For each $\theta$ in a neighborhood of $\theta^*$ there is a unique interior maximizer $\alpha^!(\theta) \in \Lambda$ of $\alpha \mapsto \mathrm{HD}^2(\theta, \alpha)$, with $\alpha^!(\theta^*) = \alpha^*$.*
*(ii) The block Hessian $\nabla^2_{\alpha\alpha} \mathrm{HD}^2(\theta^*, \alpha^*)$ is nonsingular.*
*(iii) The profile Hessian*
$$\widetilde{H}_{\theta\theta} := \nabla^2_{\theta\theta} \mathrm{HD}^2(\theta^*, \alpha^*) - \nabla^2_{\theta\alpha} \mathrm{HD}^2(\theta^*, \alpha^*) \left[\nabla^2_{\alpha\alpha} \mathrm{HD}^2(\theta^*, \alpha^*)\right]^{-1} \nabla^2_{\alpha\theta} \mathrm{HD}^2(\theta^*, \alpha^*)$$
*is positive definite.*

**Lemma S4.3** (Envelope and implicit function identities). *Assume $(H_D)$, $(H_G)$ and Assumption S4.2. Then $\alpha(\cdot)$ is $C^1$ in a neighborhood of $\theta^*$ and*

$$\nabla_\theta S(\theta) = \nabla_\theta \operatorname{HD}^2\big(\theta, \alpha(\theta)\big), \qquad \frac{d\,\alpha(\theta)}{d\theta} = -\big[\nabla^2_{\alpha\alpha} \operatorname{HD}^2\big]^{-1} \nabla^2_{\alpha\theta} \operatorname{HD}^2,$$

*evaluated at $(\theta, \alpha(\theta))$. Moreover*

$$\nabla^2_{\theta\theta} S(\theta^*) = \widetilde{H}_{\theta\theta}.$$

*Proof.* The map $(\theta, \alpha) \mapsto \operatorname{HD}^2(\theta, \alpha)$ is $C^2$ by $(H_D)$ and $(H_G)$. By Assumption S4.2(i)–(ii), we have that $\nabla_\alpha \operatorname{HD}^2(\theta, \alpha(\theta)) = 0$ and $\nabla^2_{\alpha\alpha} \operatorname{HD}^2(\theta^*, \alpha^*)$ is invertible, so the implicit function theorem gives $C^1$-smoothness of $\alpha^!(\cdot)$ and the derivative formula. The envelope identity then follows by the chain rule together with $\nabla_\alpha \operatorname{HD}^2(\theta, \alpha(\theta)) = 0$. Differentiating once more and substituting $d\alpha/d\theta$ yields the Schur-complement expression stated for $\nabla^2_{\theta\theta} S(\theta^*)$. □

**Lemma S4.4** (First-order expansion of the empirical profile score). *Assume $(H_D)$, $(H_G)$, $(H_n)$, $(H_1)$–$(H_3)$ and Assumption S4.2. Let $\alpha_n(\theta) \in \arg\max_{\alpha \in \Lambda} \operatorname{HD}^2_n(\theta, \alpha)$. Then*

$$\sqrt{n}\,\nabla_\theta S_n(\theta^*) =$$
$$\sqrt{n}\Big\{\nabla_\theta \operatorname{HD}^2_n(\theta^*, \alpha^*) - \nabla^2_{\theta\alpha} \operatorname{HD}^2(\theta^*, \alpha^*) \big[\nabla^2_{\alpha\alpha} \operatorname{HD}^2(\theta^*, \alpha^*)\big]^{-1} \nabla_\alpha \operatorname{HD}^2_n(\theta^*, \alpha^*)\Big\} + o_p(1).$$

*Consequently,*

$$\sqrt{n}\,\nabla_\theta S_n(\theta^*) \xrightarrow{d} \mathcal{N}(0,\ S_{\text{eff}}),$$

*where*

$$S_{\text{eff}} := Var\Big(\Delta_\theta - \nabla^2_{\theta\alpha} \operatorname{HD}^2 \big[\nabla^2_{\alpha\alpha} \operatorname{HD}^2\big]^{-1} \Delta_\alpha\Big),$$

$$\Delta_\theta := \sqrt{n}\,\nabla_\theta \operatorname{HD}^2_n(\theta^*, \alpha^*), \ \Delta_\alpha := \sqrt{n}\,\nabla_\alpha \operatorname{HD}^2_n(\theta^*, \alpha^*).$$

*All derivatives above are evaluated at $(\theta^*, \alpha^*)$.*

*Proof.* By Danskin's/envelope theorem in the $C^1$ case,

$$\nabla_\theta S_n(\theta) = \nabla_\theta \operatorname{HD}^2_n(\theta, \alpha_n(\theta))$$

because $\nabla_\alpha \operatorname{HD}^2_n(\theta, \alpha_n(\theta)) = 0$. Evaluating at $\theta^*$ and Taylor expanding in $\alpha$ around $\alpha^*$,

$$\nabla_\theta S_n(\theta^*) = \nabla_\theta \operatorname{HD}^2_n(\theta^*, \alpha^*) + \nabla^2_{\theta\alpha} \operatorname{HD}^2(\theta^*, \alpha^*)\,[\alpha_n(\theta^*) - \alpha^*] + r_n,$$

with $r_n = o_p(n^{-1/2})$. By standard M-estimation theory under $(H_1)$–$(H_3)$ (see, e.g., van der Vaart, 1998, Thm. 5.41), the maximizer $\alpha_n(\theta^*)$ satisfies

$$\sqrt{n}\big(\alpha_n(\theta^*) - \alpha^*\big) = -H_{\alpha\alpha}^{-1}\sqrt{n}\,\nabla_\alpha HD^2_n(\theta^*, \alpha^*) + o_p(1),$$

where $H_{\alpha\alpha} = \nabla^2_{\alpha\alpha} HD^2(\theta^*, \alpha^*)$. Next, expand the first order condition of $\alpha$ at $(\theta^*, \alpha_n(\theta^*))$:

$$0 = \nabla_\alpha \operatorname{HD}^2_n(\theta^*, \alpha_n(\theta^*)) = \nabla_\alpha \operatorname{HD}^2_n(\theta^*, \alpha^*) + \nabla^2_{\alpha\alpha} \operatorname{HD}^2(\theta^*, \alpha^*)\,[\alpha_n(\theta^*) - \alpha^*] + o_p(n^{-1/2}).$$

Since $\nabla^2_{\alpha\alpha} \operatorname{HD}^2(\theta^*, \alpha^*)$ is nonsingular,

$$\alpha_n(\theta^*) - \alpha^* = -\big[\nabla^2_{\alpha\alpha} \operatorname{HD}^2(\theta^*, \alpha^*)\big]^{-1} \nabla_\alpha \operatorname{HD}^2_n(\theta^*, \alpha^*) + o_p(n^{-1/2}).$$

Substitute into the first expansion, multiply by $\sqrt{n}$, and use Proposition 3.1 to obtain the stated linear representation and asymptotic normality with covariance $S_{\text{eff}}$. □

The next lemma is concerned with the convergence of the empirical profile Hessian.

**Lemma S4.5.** *Under the assumptions of Lemma S4.4,*
$$\nabla^2_{\theta\theta} S_n(\tilde{\theta}_n) \xrightarrow{p} \widetilde{H}_{\theta\theta}$$
*for any sequence $\tilde{\theta}_n \to \theta^*$.*

*Proof.* Let $S_n(\theta) = \mathrm{HD}_n^2(\theta, \alpha_n(\theta))$ and differentiate, by the chain rule,
$$\nabla^2_{\theta\theta} S_n(\theta) = \nabla^2_{\theta\theta} \mathrm{HD}_n^2(\theta, \alpha_n(\theta)) + \nabla^2_{\theta\alpha} \mathrm{HD}_n^2(\theta, \alpha_n(\theta)) \frac{d\,\alpha_n(\theta)}{d\theta}.$$

Differentiating the empirical first order condition of $\alpha$, $\nabla_\alpha \mathrm{HD}_n^2(\theta, \alpha_n(\theta)) = 0$ yields
$$\frac{d\,\alpha_n(\theta)}{d\theta} = -\left[\nabla^2_{\alpha\alpha} \mathrm{HD}_n^2(\theta, \alpha_n(\theta))\right]^{-1} \nabla^2_{\alpha\theta} \mathrm{HD}_n^2(\theta, \alpha_n(\theta))$$

whenever the inverse exists. By Proposition 3.2 (elementwise convergence of second derivatives) and consistency $\alpha_n(\tilde{\theta}_n) \to \alpha^*$, each empirical block converges in probability to its population counterpart, hence the whole expression converges to the Schur complement $\widetilde{H}_{\theta\theta}$. □

**Theorem S4.2** (CLT for the profiled generator estimator). *Assume $(H_D)$, $(H_G)$, $(H_n)$, $(H_1)$–$(H_3)$ and Assumption S4.2. Let $\hat{\theta}_n \in \arg\min_{\theta \in \Theta} S_n(\theta)$ and suppose $\Theta$ is compact and $S$ has a unique minimizer $\theta^*$. Then*
$$\sqrt{n}\,(\hat{\theta}_n - \theta^*) \xrightarrow{d} \mathcal{N}\!\left(0,\ \widetilde{H}_{\theta\theta}^{-1}\, S_{\mathrm{eff}}\, \widetilde{H}_{\theta\theta}^{-1}\right),$$
*with $\widetilde{H}_{\theta\theta}$ and $S_{\mathrm{eff}}$ as in Lemmas S4.3 and S4.4.*

*Proof.* By optimality, $0 = \nabla_\theta S_n(\hat{\theta}_n) = \nabla_\theta S_n(\theta^*) + \nabla^2_{\theta\theta} S_n(\tilde{\theta}_n)\,(\hat{\theta}_n - \theta^*)$ for some $\tilde{\theta}_n$ on the line segment between $\hat{\theta}_n$ and $\theta^*$. Multiply by $\sqrt{n}$ and rearrange:
$$\sqrt{n}\,(\hat{\theta}_n - \theta^*) = -\left[\nabla^2_{\theta\theta} S_n(\tilde{\theta}_n)\right]^{-1} \sqrt{n}\,\nabla_\theta S_n(\theta^*).$$

By Lemma S4.5, $\nabla^2_{\theta\theta} S_n(\tilde{\theta}_n) \xrightarrow{p} \widetilde{H}_{\theta\theta}$, which is invertible by Assumption S4.2(iii), and by Lemma S4.4, $\sqrt{n}\,\nabla_\theta S_n(\theta^*) \xrightarrow{d} \mathcal{N}(0, S_{\mathrm{eff}})$. The result follows by Slutsky's theorem. □

**Remark S4.2** (Connection with Theorem 3.3). Let $H := \nabla^2 \mathrm{HD}^2(\theta^*, \alpha^*)$ and write it in block form with respect to $(\alpha, \theta)$. The joint CLT of Theorem 3.3 gives $\sqrt{n}\big((\theta_n, \alpha_n) - (\theta^*, \alpha^*)\big) \xrightarrow{d} \mathcal{N}(0, \Sigma)$, where $\Sigma = H^{-1} S (H^{-1})^\top$ with $S = Var(\sqrt{n}\,\nabla \mathrm{HD}_n^2)$. A standard block inversion shows that the $\theta$-marginal covariance equals $\widetilde{H}_{\theta\theta}^{-1} S_{\mathrm{eff}} \widetilde{H}_{\theta\theta}^{-1}$, i.e. it coincides with the variance in Theorem S4.2. Thus the profiled CLT is exactly the $\theta$-component of the joint CLT.

**Remark S4.3** (Approximated objective). All statements above hold verbatim with $\mathrm{HD}_n^2$ replaced by $\widetilde{L}_n$ (Section 2.1), since the derivative blocks and limit arguments used in Lemmas S4.3–S4.5 are the same.

Table S1: Median (standard deviation) of MSE (×100) of $\mu$ at the final epoch for the considered methods and percentage of contamination $\varepsilon$.

|  | $\varepsilon$=0% | $\varepsilon$=1% | $\varepsilon$=5% | $\varepsilon$=10% | $\varepsilon$=20% |
|---|---|---|---|---|---|
| GAN | **0.01** (1.9) | **0.15** (11.6) | **3.28** (1046.8) | 54.89 (2124.0) | 152.37 (490.5) |
| WGAN | 34.27 (15.8) | 25.03 (31.0) | 116.46 (4283.5) | 230.51 (1949.5) | 578.34 (4954.6) |
| Approx. HD | 0.99 (7.7) | 0.73 (385.7) | 21.96 (921.5) | **2.67** (291.7) | 146.72 (129.0) |
| HD ($c_n$=0.0001) | 63.33 (68.4) | 18.78 (6201.4) | 11.00 (3734.6) | 15.12 (2137.7) | **7.22** (374.9) |
| HD ($c_n$=0.01) | 26.69 (1072.7) | 32.72 (7199.9) | 83.15 (4426.2) | 37.21 (5805.2) | 147.95 (4804.0) |
| HD ($c_n$=0.5) | 25.66 (1358.5) | 20.29 (128.6) | 17.75 (691.6) | 93.31 (3533.9) | 100.51 (514.8) |

Table S2: Median (standard deviation) of MSE (×100) of $\sigma$ at the final epoch for the considered methods and percentage of contamination $\varepsilon$.

|  | $\varepsilon$=0% | $\varepsilon$=1% | $\varepsilon$=5% | $\varepsilon$=10% | $\varepsilon$=20% |
|---|---|---|---|---|---|
| GAN | 11.18 (94.8) | 3.06 (220.3) | **28.52** (595.4) | 331.04 (1939.5) | 446.08 (836.9) |
| WGAN | 5.00 (71.6) | **2.18** (14.7) | 69.05 (2662.6) | 85.24 (2427.0) | 1110.06 (2078.6) |
| Approx. HD | **0.58** (58.5) | 30.40 (209.5) | 827.90 (708.0) | **9.19** (619.8) | 255.01 (665.0) |
| HD ($c_n$=0.0001) | 10.31 (27.2) | 20.93 (1312.4) | 33.62 (1538.4) | 37.02 (914.8) | **29.62** (1096.0) |
| HD ($c_n$=0.01) | 16.07 (55.1) | 39.62 (1005.6) | 112.32 (1313.8) | 184.94 (2926.0) | 675.86 (8350.9) |
| HD ($c_n$=0.5) | 28.17 (2796.4) | 30.00 (39.9) | 65.55 (2393.2) | 243.55 (1306.6) | 584.61 (955.6) |

Table S3: Median (standard deviation) of RMSE (×100) for $(\mu, \sigma)$ at the final epoch for the considered methods and percentage of contamination $\varepsilon$.

|  | $\varepsilon$=0% | $\varepsilon$=1% | $\varepsilon$=5% | $\varepsilon$=10% | $\varepsilon$=20% |
|---|---|---|---|---|---|
| GAN | 23.51 (43.5) | **12.30** (68.8) | **41.11** (134.6) | 171.53 (206.3) | 173.59 (141.4) |
| WGAN | 44.20 (16.9) | 36.71 (15.2) | 105.47 (291.5) | 125.57 (231.6) | 351.97 (304.7) |
| Approx. HD | **11.80** (28.1) | 43.31 (85.9) | 213.45 (138.3) | **23.86** (124.0) | 140.27 (114.6) |
| HD ($c_n$=0.0001) | 65.24 (26.6) | 57.23 (259.1) | 55.11 (246.9) | 47.47 (173.7) | **42.98** (146.3) |
| HD ($c_n$=0.01) | 54.12 (85.3) | 65.00 (232.5) | 107.42 (173.9) | 110.96 (214.1) | 202.90 (281.9) |
| HD ($c_n$=0.5) | 54.58 (154.1) | 50.09 (31.3) | 65.56 (139.2) | 131.67 (169.0) | 190.86 (85.5) |

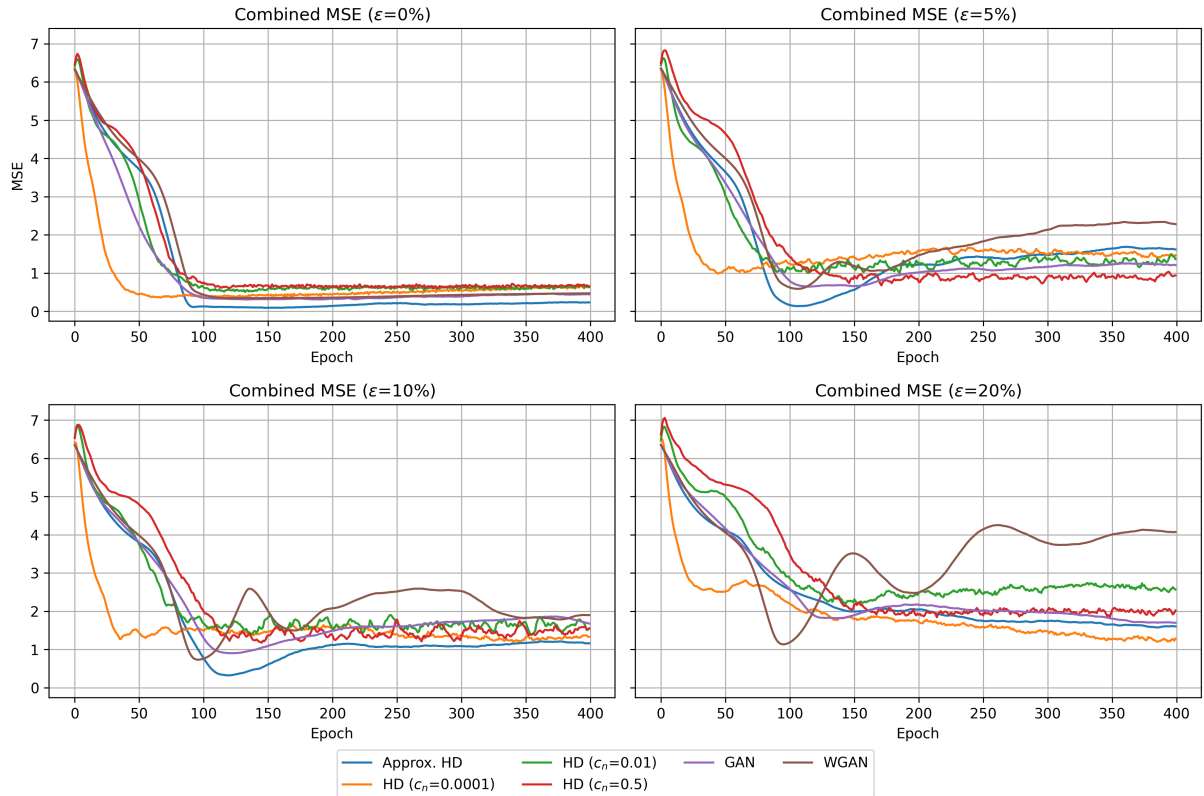

Figure S1: Mean Squared Error (MSE) per epoch for the different contamination percentage $\varepsilon$ comparing the different methods.

## S5 Additional Simulation Results

In this section we complement the simulation study results presented in Section 5 of the manuscript. Figure S2 reports, for each contamination proportion $\varepsilon \in \{0\%, 5\%, 10\%, 20\%\}$, the RMSE per epoch of the estimated location–scale parameters in the Gaussian model for all considered loss functions, namely the standard GAN, WGAN, the approximate Hellinger loss, and the KDE-based Hellinger losses with different bandwidth choices. Tables S1 and Table S2 report the median and standard deviation of the MSE for the location and scale parameters of the competing estimators at the final training epoch. While Table S3 shows the RMSE at the final epoch.

## References

N. Glick. Consistency conditions for probability estimators and integrals of density estimators. *Utilitas Mathematica*, 6:64 – 74, 1974.



\end{document}